%% file: arxiv.tex
\documentclass{article}

\usepackage{geometry}
\input{def}
\usepackage{amsthm}
\renewcommand{\updated}[1]{#1}

\newtheorem{example}{Example}

\newtheorem{theorem}{Theorem}
\newtheorem{corollary}{Corollary}
\newtheorem{lemma}{Lemma}
\newtheorem{proposition}{Proposition}

\title{Optimal Clustering from Noisy Binary Feedback}

\author{
    Kaito Ariu\thanks{School of Electrical Engineering and Computer Science, KTH. Also with CyberAgent.}
    \and
    Jungseul Ok\thanks{Department of Computer Science and Engineering, POSTECH.}
    \and
    Alexandre Proutiere\thanks{School of Electrical Engineering and Computer Science, KTH.}
    \and
    Se-Young Yun\thanks{Graduate School of AI, KAIST.}
}

\begin{document}

\maketitle
\begin{abstract}

\input{abstract}

\end{abstract}
\input{intro}

\input{lower_bounds}

\input{algorithms}

\input{Numerical_experiments}

\input{nonsynthetic}

\input{Conclusion}

\bibliographystyle{plain}
\bibliography{References}

\newpage
\appendix

\input{app_table}

\input{appendix}

\end{document}

%% file: def.tex
\usepackage{amsmath}%
\usepackage{amssymb}
\usepackage{bbm}
\usepackage{paralist}
\usepackage{enumitem}
\usepackage{algorithm}
\usepackage{algorithmic}
\usepackage{xcolor}
\usepackage{mathtools}
\usepackage[switch]{lineno}
\usepackage{dirtytalk}
\usepackage{booktabs}
\usepackage{url}
\usepackage{subcaption}

\newcommand{\ep}{\hfill $\Box$}

\renewcommand{\vec}{\boldsymbol}

\DeclareMathOperator{\KL}{\textnormal{KL}}

\DeclareMathOperator{\EXP}{\mathbb{E}}
\newcommand{\qand}{\quad \text{and} \quad}

\renewcommand{\tilde}{\widetilde}

\renewcommand{\Pr}{\mathbb{P}}
\renewcommand{\star}{*}
\newcommand{\set}[1]{\mathcal{#1}}

\newcommand{\kl}{\textnormal{KL}}

\DeclareMathOperator*{\argmin}{arg\,min}
\DeclareMathOperator*{\argmax}{arg\,max}

\newcommand{\updated}[1]{\textcolor{red}{#1}}

%% file: abstract.tex
We study the problem of clustering a set of items from binary user feedback. Such a problem arises in crowdsourcing platforms solving large-scale labeling tasks with minimal effort put on the users. For example, in some of the recent reCAPTCHA systems, users clicks (binary answers) can be used to efficiently label images. In our inference problem, items are grouped into initially unknown non-overlapping clusters. To recover these clusters, the learner sequentially presents to users a finite list of items together with a question with a binary answer selected from a fixed finite set. For each of these items, the user provides a noisy answer whose expectation is determined by the item cluster and the question and by an item-specific parameter characterizing the {\it hardness} of classifying the item. The objective is to devise an algorithm with a minimal cluster recovery error rate. We derive problem-specific information-theoretical lower bounds on the error rate satisfied by any algorithm, for both uniform and adaptive (list, question) selection strategies. For uniform selection, we present a simple algorithm built upon the K-means algorithm and whose performance almost matches the fundamental limits. For adaptive selection, we develop an adaptive algorithm that is inspired by the derivation of the information-theoretical error lower bounds, and in turn allocates the budget in an efficient way. The algorithm learns to select items hard to cluster and relevant questions more often. We compare the performance of our algorithms with or without the adaptive selection strategy numerically and illustrate the gain achieved by being adaptive.

%% file: intro.tex
\section{Introduction}

Modern Machine Learning (ML) models require a massive amount of labeled data to be efficiently trained. Humans have been so far the main source of labeled data. This data collection is often tedious and very costly. Fortunately, most of the data can be simply labeled by non-experts. This observation is at the core of many crowdsourcing platforms such as reCAPTCHA, where users receive low or no payment. In these platforms, complex labeling problems are decomposed into simpler tasks, typically questions with binary answers. In reCAPTCHAs, for example, the user is asked to click on images (presented in  batches) that contain a particular object (a car, a road sign), and the system leverages users' answers to label images. As another motivating example, consider the task of classifying bird images. Users may be asked to answer binary questions like: \say{Is the bird grey?}, \say{Does it have a circular tail fin?}, \say{Does it have a pattern on its cheeks?}, etc. Correct answers to those questions, if well-processed, may lead to an accurate bird classification and image labels. In both aforementioned examples, some images may be harder to label than others, e.g., due to the photographic environment, the birds' posture, etc. Some questions may be harder to answer than others, leading to a higher error rate. To build a reliable system, tasks/questions have to be carefully designed and selected, and user responses need to be smartly processed. Efficient systems must also learn the difficulty of the different tasks, and guess how informative they are when solving the complex labeling problem.

\updated{This paper investigates the design of such systems, tackling clustering problems that have to be solved using answers to binary questions. We incorporate a model that takes into consideration the varying difficulty levels or {\it heterogeneity} of clustering each item.}
We propose a full analysis of the problem, including information-theoretical limits that hold for {\it any} algorithm and novel algorithms with provable performance guarantees. Before giving a precise statement of our results, we provide a precise description of \updated{the problem setting and} the statistical model dictating the way users answer. This model is inspired by models, such as the Dawid-Skene model \cite{dawid1979maximum}
successfully used in the crowdsourcing literature, see e.g., \cite{oh2016} and references therein. However, to the best of our knowledge, this paper is the first to model and analyze clustering problems with binary feedback and accounting for item heterogeneity.

\subsection{\updated{Problem setting and feedback model}}

\updated{Consider a large set $\set{I}$ of $n$ items (e.g. images) partitioned into $K$ disjoint unknown clusters $\set{I}_1,\ldots, \set{I}_K$. Denote by $\sigma(i)$ the cluster of item $i$. To recover these hidden clusters, the learner gathers binary user feedback sequentially. Upon arrival, a user is presented a list of $w\ge 1$ items together with a question with a binary answer. The question is selected from a predefined finite set of cardinality $L$. The process of selecting the (list, question) pair for a given user can be carried out in either a nonadaptive or adaptive manner (in the latter case, the pair would depend on user feedback previously collected). Importantly, our model captures item heterogeneity: the difficulty of clustering items varies across items. We wish to devise algorithms recovering clusters as accurately as possible using the noisy binary answers collected from $T$ users. }

We use the following statistical model parametrized by a matrix $\vec{p}:=(p_{k\ell})_{k \in [K], \ell \in [L]}$\footnote{Define for any integer $A\ge 1$, the set $[A]:=\{1,\ldots,A\}$.
} with entries in $[0,1]$ and by a vector $\vec{h}:=(h_i)_{i \in \set{I}}\in [1/2,1]^n$. These parameters are (initially) unknown. When the $t$-th user is asked a question $\ell_t =\ell \in [L]$ for a set $\set{W}_t$ of $w \ge 1$ items, she provides noisy answers: for the item $i\in \set{I}_k$ in the list $\set{W}_t$, her answer $X_{i \ell t}$ is $+1$ with probability 
$q_{i\ell} := h_i p_{k \ell }+\bar{h}_i\bar{p}_{k \ell}$, and $-1$ with probability $\bar{q}_{i \ell}$, where
for brevity, $\bar{x}$ denotes $1-x$ for any $x \in [0, 1]$. Answers are independent across items and users. Our model is simple, but general enough to include as specific cases, crowdsourcing models recently investigated in the literature.  For example, the model in \cite{oh2016} corresponds to our model with only one question ($L=1$), two clusters ($K=2$), and a question asked for a single item at a time ($w=1$). Note that in our model, answers are collected from a very large set of users, and a given user is very unlikely to interact with the system several times. This justifies the fact that answers provided by the various users are statistically identical. 

\medskip
\noindent
{\bf Item hardness.} An important aspect of our model stems from the item-specific parameter $h_i$. It can be interpreted as the {\it hardness} of clustering item $i$, whereas $p_{k \ell}$  corresponds to a latent parameter related to question $\ell$ when asked for an item in cluster $k$. Note that 
when $h_i=1/2$, $q_{i \ell} = {1/2}$ irrespective of the cluster of item $i$. Hence any question $\ell$ on item $i$ receives completely random responses, and this item cannot be clustered. Further note that intuitively, the larger the hardness parameter $h_i$ of item $i$ is, the easier it can be clustered. Indeed, when asking question $\ell$, we can easily distinguish whether item $i$ belongs to cluster $k$ or $k'$ if the difference between the corresponding parameters of user statistical answers $h_i p_{k \ell }+\bar{h}_i\bar{p}_{k \ell}$ and $h_i p_{k' \ell }+\bar{h}_i\bar{p}_{k' \ell}$ is large. This difference is $|  p_{k \ell }-p_{k' \ell }|(2h_i-1)$, an increasing function of $h_i$. We believe that introducing item heterogeneity is critical to obtain a realistic model (without $\vec{h}$, all items from the same cluster would be exchangeable), but complicates the analysis. Most theoretical results on clustering or community detection do not account for this heterogeneity -- refer to Section \ref{section:related} for detail.

\medskip
\noindent
\updated{{\bf Illustrative Example.} We introduce an example to illustrate the structure and characteristics of our model.}

\begin{figure}[h!]
\begin{subfigure}{0.5\linewidth}
  \centering\includegraphics[width=0.9\textwidth]{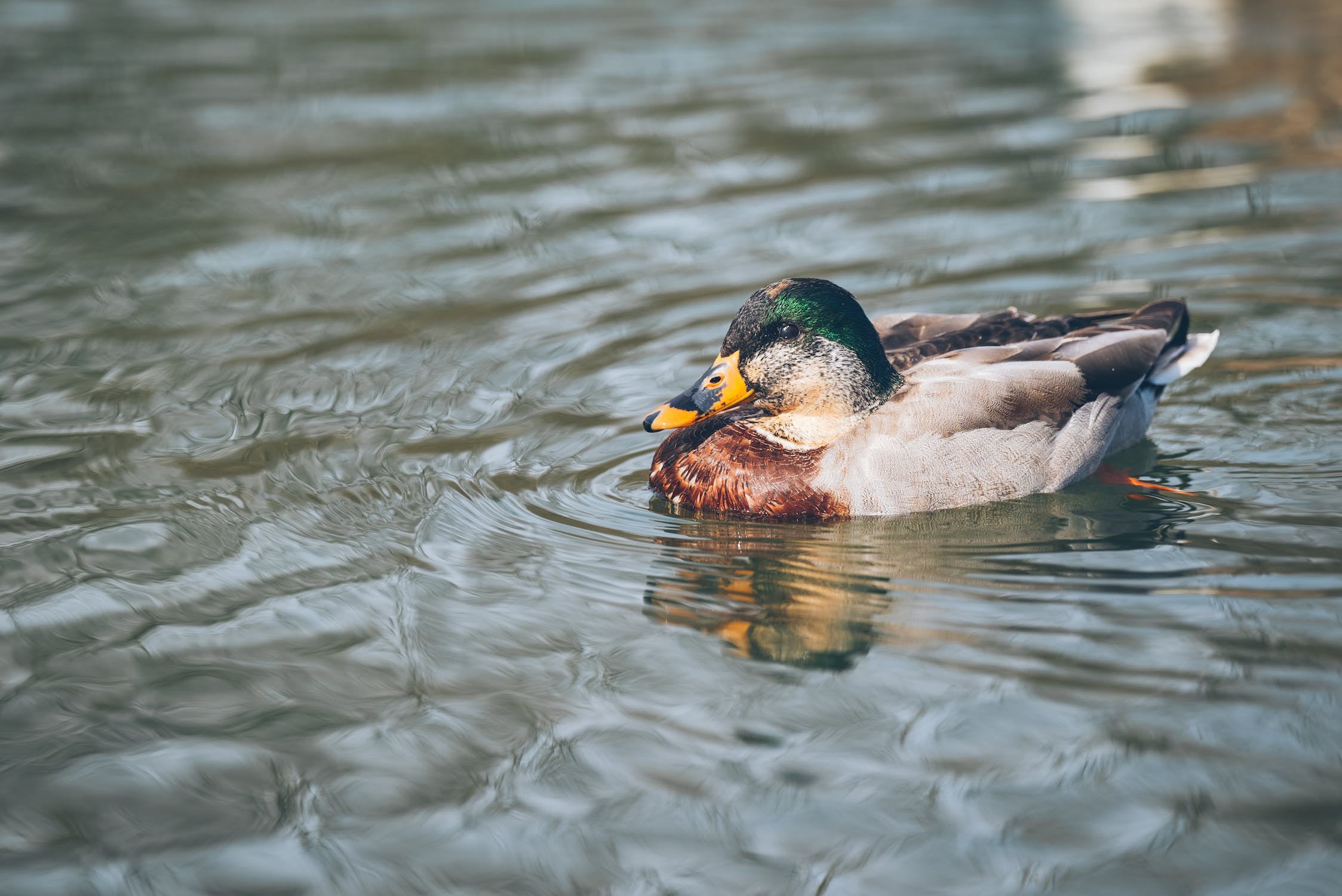}
  \caption{\updated{ Mallard \cite{mallard2024}}}
  \label{fig:mallard}
  \end{subfigure}
  \begin{subfigure}{0.5\linewidth}
  \includegraphics[width=0.9\textwidth]{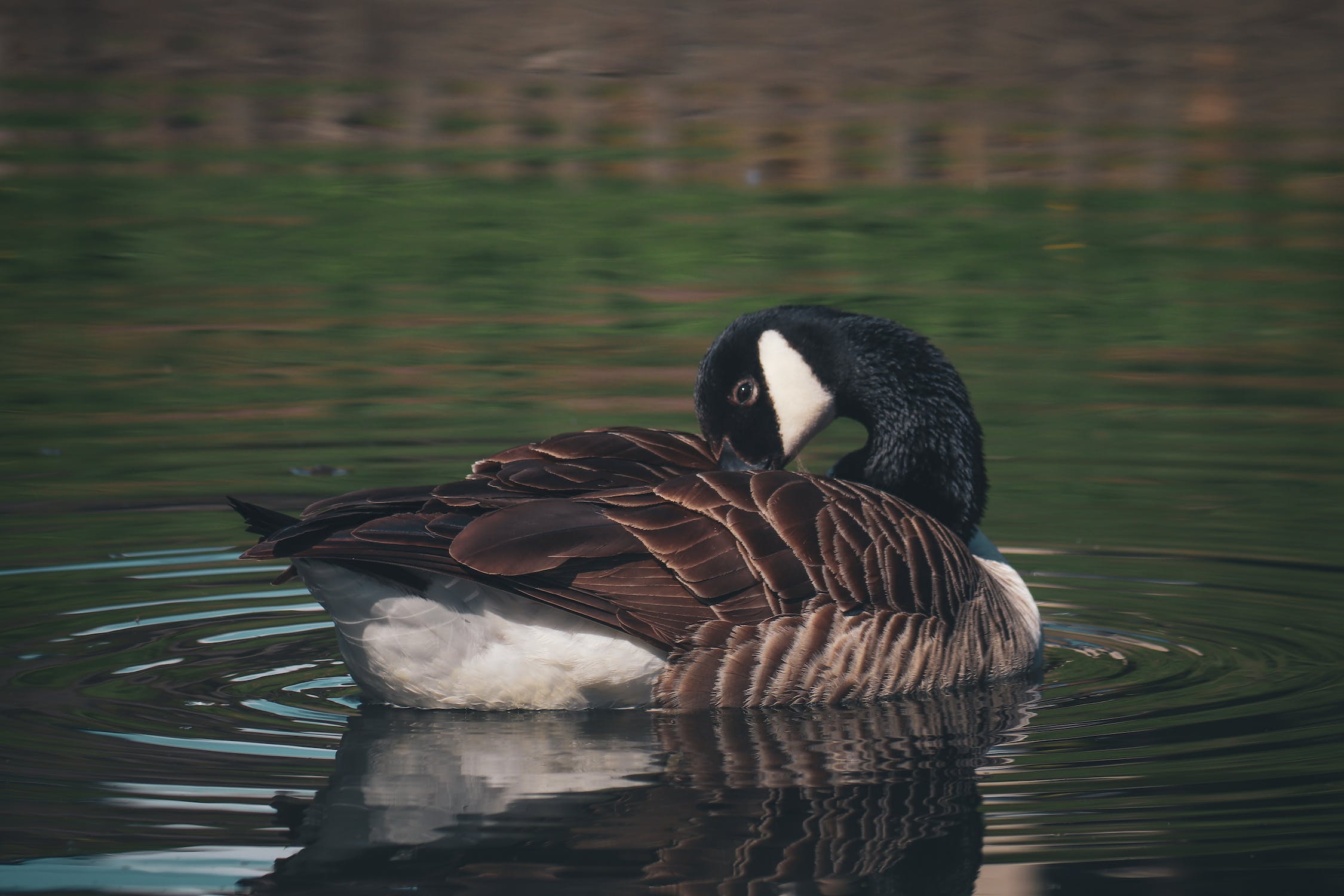}
  \caption{\updated{ Canadian Goose \cite{geese2024}}}
  \label{fig:goose}
  \end{subfigure}

\caption{\updated{Images for Example 1.}}
\end{figure}

\updated{
\begin{example}\label{ex:1}
    Consider the task of classifying images into two types of birds: Mallards and Canadian Geese. Mallards (see Figure~\ref{fig:mallard} for an image), a type of duck, and Canadian Geese (see Figure~\ref{fig:goose} for an image), which are not classified as ducks, present a unique classification challenge. In this case, $L=1$, and the question posed to the users is: ``Is the bird in the image a duck?". We assign cluster $1$ for the Mallard images and cluster $2$ for the Canadian Goose images. Assume that $p_{11}=0.8$ and  $p_{21}=0.3$: they are latent probabilities of answering yes to the question given an image of a Mallard 
    and a Canadian Goose, respectively.
\end{example}
}
\noindent
\updated{$p_{11}$ and $p_{21}$ represent the latent probabilities of answering yes to a question $1$. These parameters also consider the scenario where a user, randomly selected from a large set, may not answer a question correctly due to a lack of knowledge or other reasons. For each image $i$, $h_i$ indicates the difficulty of classification. For instance, when the image is of Mallards (a type of duck), and the image is clear, the classification is relatively easy, and $h_i$ is set to $h_i=1$. Consequently, the probability of correctly identifying the Mallards is $q_{i1} = h_ip_{11} + \bar{h}_i\bar{p}_{11} = 1\cdot 0.8 + 0\cdot0.2=0.8$. However, when another image $j$ of Mallards is blurred due to poor lighting or other factors, and the classification difficulty is $h_j=0.8$, the probability of correct classification decreases to $ q_{j1} = h_jp_{11} + \bar{h}_j\bar{p}_{11} = 0.8\cdot 0.8 + 0.2\cdot0.2=0.68$. As a result, the feedback obtained for image $j$ is more ambiguous compared to that for image $i$, due to the increased difficulty in classification.}

\medskip
\noindent
{\bf Assumptions.} We make the following mild assumptions on our statistical model $\set{M} := (\vec{p}, \vec{h})$. Define for each $k \in [K]$, $\vec{r}_k := (r_{k \ell})_{\ell \in [L]}$ with  ${r}_{k\ell} := {2p_{k\ell} - 1}$. Throughout the paper, $\|\cdot\|$ denotes the $\ell_\infty$-norm, i.e., $\|\vec{x}\| = \max_i{|x_i|}$.
\begin{linenomath}
\begin{align*}
\begin{array}{rlrl}
\text{(A1)} &  h_{\star} := \min_{i \in \set{I}} (2h_i -1)\in (0,1),& 
\quad\text{(A2)} &  \exists \eta>0, 
\eta \le p_{k \ell} \le 1-\eta.
\\
& \rho_{\star} := \min_{k \neq k'}\min_{\updated{c\ge0}} \|c \vec{r}_{k'} - \vec{r}_k \| > 0 . & & 
\end{array}
\end{align*}
\end{linenomath} 
Assumption (A1) excludes the cases where clustering is impossible even if all parameters were accurately estimated. Indeed, when $h_\star = 0$, there exists at least one item $i$ which receives completely random responses for any question, i.e., $q_{i\ell} = 1/2$ for any $\ell \in [L]$. 
\updated{Observe that when $\rho_\star  = 0$, 
there exists $k \neq k'$ and $c \ge 0$ such that $2{p}_{k \ell} -1 = c (2 {p}_{k' \ell} -1)$
for all $\ell \in [L]$.
Then, for item $i \in \set{I}_k$, %
we find $h' \in [1/2,1]$ such that $2q_{i\ell} -1 = (2h' -1)(2p_{k'\ell} - 1)$.  Items in the different clusters $k$ and $k'$ can have the same value of $q_{i \ell}$. As a consequence, from the answers, we cannot determine whether $i$ is in cluster $k$ or $k'$. In Example~\ref{ex:1}, $r_{11} = 0.6$, $r_{21}= -0.4$,  and the value of $\rho_*$ is $\rho_* = |0 \cdot 0.6 + 0.4 |= 0.4$. } Assumption (A2) states some homogeneity among the parameters of the clusters.
It implies that $q_{i \ell} \in [\eta, 1- \eta]$ for all $i \in \set{I}$ and $ \ell \in [L]$. Let $\Omega$ be the set of all models satisfying (A1) and (A2). 

\updated{For convenience, we provide a table summarizing all the notations in Appendix~A.}

\subsection{Main contributions} 

We study both nonadaptive and adaptive sequential (list, question) selection strategies. In the case of nonadaptive strategy, we assume that the selection of (list, question) pairs is {\it uniform} in the sense that the number of times a given question is asked for a given item is (roughly) $\lfloor Tw/(nL)\rfloor$. The objective is to devise a clustering algorithm taking as input the data collected over $T$ users and returning estimated clusters as accurate as possible.  When using adaptive strategies, the objective is to devise an algorithm that sequentially selects the (list, question) pairs presented to users, and that, after having collected answers from $T$ users, returns accurate estimated clusters.  

Our contributions are as follows. We first derive information-theoretical performance limits satisfied by any algorithm under uniform or adaptive sequential (list, question) selection strategy. We then propose a clustering algorithm that matches our limits order-wise in the case of uniform (list, question) selection. We further present a joint adaptive (list, question) selection strategy and clustering algorithm, and illustrate, using numerical experiments on both synthetic and real data, the advantage of being adaptive. 

\medskip
\noindent
{\bf Fundamental limits.} We provide a precise statement of our lower bounds on the cluster recovery error rate. These bounds are problem specific, i.e., they depend explicitly on the model $\set{M} = (\vec{p}, \vec{h})$, and they will guide us in the design of algorithms.

\medskip
{\it (Uniform selection)} In this case, we derive a clustering error lower bound for each individual item. Let $\pi$ denote a clustering algorithm, and define the clustering error rate of item $i \in \set{I}$ as $\varepsilon^\pi_i (n, T):=\Pr [i \in \set{E}^\pi]$, where $\set{E}^\pi$ denotes the set of mis-classified items under $\pi$. The latter set is defined as $\set{E}^\pi:= \cup_{k\in [K]} \set{I}_k\setminus \set{S}_{\gamma(k)}^\pi$, where $(\set{S}_{1}^\pi,\ldots, \set{S}_{K}^\pi)$ denotes the output of $\pi$ and $\gamma$ is a permutation of $[K]$ minimizing the cardinality of $\cup_{k\in [K]} \set{I}_k\setminus \set{S}_{\zeta(k)}^\pi$ over all possible permutations $\zeta$ of $[K]$. When deriving problem-specific error lower bounds, we restrict our attention to so-called {\it uniformly good} algorithms. An algorithm $\pi$ is {\it uniformly good} if for all $\set{M} \in \Omega$ and $i \in \set{I}$, $\varepsilon^\pi_i (n, T)= o(1)$ as $T \to \infty$ under $T= \omega(n)$. We establish that for any $\set{M}\in \Omega$ satisfying (A1) and (A2), under any uniformly good clustering algorithm $\pi$, as $T$ grows large under $T=\omega(n)$, for any item $i$, we have:
\begin{align}
&\varepsilon^\pi_i(n, T)  \ge  \exp \left( - \frac{Tw}{n} \set{D}_{\set{M}}^U(i) (1+ o(1)) \right), \\
&\hbox{where } \ \ \ \set{D}_{\set{M}}^U(i) := 
 \min_{k' \neq \sigma(i)}
\min_{h' \in [(h_* + 1)/2, 1]}
\frac{1}{L}\sum_{\ell} \KL(h'p_{k'\ell} + \bar{h}'\bar{p}_{k'\ell}, q_{i\ell}) > 0. 
\end{align}
In the above definition of the {\it divergence} $\set{D}_{\set{M}}^U(i)$, $\KL (a, b)$ is the Kullback-Leibler divergence between two Bernoulli distributions of means $a$ and $b$ ($\KL (a, b) := a \log \frac{a}{b}+\bar{a} \log \frac{\bar{a}}{\bar{b}}$). Note that uniformly good algorithms actually exist (see Algorithm 1 presented in Section \ref{section4}).

\medskip
{\it (Adaptive selection)} We also provide clustering error lower bounds in the case the algorithm is also sequentially selecting (list, question) pairs in an adaptive manner. Note that here a lower bound cannot be derived for each item individually, say item $i$, since an adaptive algorithm could well select this given item often so as to get no error when returning its cluster. Instead we provide a lower bound for the cluster recovery error rate averaged over all items, i.e., $\varepsilon^\pi(n,T):={ \frac{1}{n} }\sum_{i\in \set{I}}\varepsilon^\pi_i(n, T)$. Under any uniformly good joint (list, question) selection and clustering algorithm $\pi$, as $T$ grows large under $T = \omega(n)$, we have:
\begin{align}
&\varepsilon^\pi(n, T) \ge  \exp \left( - \frac{Tw}{n} \tilde{\set{D}}_{\set{M}}^A (1+ o(1)) \right), 
\label{eq:adapt}
\\
&\hbox{where } \ \ \ \tilde{\set{D}}_{\set{M}}^A := 
\max_{\vec{y} \in \set{Y}(n)} - \frac{n}{Tw}\log \left( \frac{1}{n} \sum_{i=1}^n \exp \left( - \frac{Tw}{n} {\set{D}}_{\set{M}}^A (i, \vec{y}) \right) \right),\label{eq:maxalloc}
\\
&\ \ \ \ \ \ \ \ \ \ \ \ \ \  {\set{D}}_{\set{M}}^A (i, \vec{y}) :=
\min_{ j : \sigma(j) \neq \sigma(i)} \sum_{\ell} \left( y_{j\ell} \KL( q_{j \ell}, q_{i\ell})+y_{i\ell} \KL( q_{i \ell}, q_{j\ell}) \right),\label{eq:divergence_adaptive}
\\
&\hbox{and } \ \ \ \set{Y}(n) :=  \left\{
\vec{y} \in [0, 1]^{n\times L} :  \sum_{i \in \set{I}, \ell \in [L]} y_{i \ell} = n
\right\}.
\end{align}
In the above lower bound, the vector $\vec{y}$ encodes the expected numbers of times the various questions are asked for each item. Specifically, as shown later, $y_{i \ell}\frac{Tw}{n}$ can be interpreted expected number of times the question $\ell$ is asked for the item $i$. Maximizing over $\vec{y}$ in (\ref{eq:maxalloc}) hence corresponds to an optimal (list, question) selection strategy, and to the minimal error rate. Further interpretations and discussions of the divergences $\set{D}_{\set{M}}^U(i)$ and ${\set{D}}_{\set{M}}^A(i, \vec{y})$ are provided later in the paper.

\medskip
\noindent
{\bf Algorithms.} We develop algorithms with both uniform and adaptive (list, question) selection strategies. 

\medskip
{\it (Uniform selection)} In this case, for each item $i$ and based on the collected answers, we build a normalized vector (of dimension $L$) that concentrates (when $T$ is large) around a vector depending on the cluster id $\sigma(i)$ only. Our algorithm applies a K-means algorithm to these vectors (with an appropriate initialization) to reconstruct the clusters. We are able to establish that the algorithm almost matches our fundamental limits. More precisely, when ${T} = \omega\left( n\right)$ and $T = o (n^2)$, under our algorithm, we have, for some absolute constant $C>0$,
\begin{linenomath}
\begin{equation}
\varepsilon_i^\pi(n,T)\leq  \exp \left( - C (2 h_i -1)^2  \rho_*^2 \frac{Tw}{Ln}(1 + o(1))\right).
\end{equation} 
\end{linenomath}
The above error rate has an optimal scaling in $T, w, L, n$. By deriving upper and lower bounds on $\set{D}_{\set{M}}^U(i)$), we further show that the scaling is also optimal in $(2h_i-1)^2$ and almost in $\rho_*$ (see Assumption (A1)). 

\medskip
{\it (Adaptive selection)} The design of our adaptive algorithm is inspired by the information-theoretical lower bounds. The algorithm periodically updates estimates of the model parameters, and of the clusters. Based on these estimates, we further estimate lower bounds on the probabilities to misclassify every item. The items we select are those with the highest lower bounds (the items that are most likely to be misclassified); we further select the question that would be the most informative about these items. We believe that our algorithm should approach the minimal possible error rate (since it follows the optimal (list, question) selection strategy). Our numerical experiments suggest that the adaptive algorithm significantly outperforms algorithms with uniform (list, question) selection strategy, especially when items have very heterogenous hardnesses.

\section{Related work}\label{section:related} 

To our knowledge, the model proposed in this paper has been neither introduced nor analyzed in previous work. The problem has similarities with crowdsourced classification problems with a very rich literature \cite{dawid1979maximum}, \cite{raykar2010learning} \cite{karger2011iterative}, \cite{zhou2012learning}, \cite{ho2013adaptive}, \cite{tran2013efficient}, \cite{zhang2014spectral}, \cite{gao2016exact}, \cite{ok2016optimality} (Dawid-Skene model and its various extensions without clustered structure), \cite{vinayak2016crowdsourced}, \cite{gomes2011crowdclustering} (Clustering without item heterogeneity).
However, our model has clear differences.
For instance, if we draw a parallel between our model and that considered in \cite{oh2016},
there tasks correspond to our items, and there are only two clusters of tasks. More importantly, the statistics of the answers for a particular task do not depend on the true cluster of the task since the ground truth is defined by the majority of answers given by the various users. 
Our results also differ from those in the crowdsourcing literature from a methodological perspective. In this literature, fundamental limits are rarely investigated, and if they are, they are in the minimax sense by postulating the worst parameter setting (e.g., \cite{zhang2014spectral}, \cite{oh2016}, \cite{gao2016exact}) or it is problem-specific but without quantifying of the error rate (e.g., \cite{ok2016optimality}). Here we derive more precise problem-specific lower bounds on the error rate, i.e., we provide minimum clustering error rates given the model parameters $(\vec{p}, \vec{h})$. Further note that most of the classification tasks studied in the literature are simple (can be solved using a single binary question).

Our problem also resembles cluster inference problems in the celebrated Stochastic Block Model (SBM), see \cite{Abbe18} for a recent survey. Plain SBM models, however, assume that the statistics of observations for items in the same cluster are identical (there are no items harder to cluster than others, this corresponds to $h_i = 1, \forall i \in \set{I}$ in our model), and observations are typically not operated in an adaptive manner. The closest work in the context of SBM to ours is the analysis of the so-called Degree-Corrected SBM, where each node is associated with an average degree quantifying the number of observations obtained for this node. The average degree then replaces our hardness parameter $h_i$ for item $i$. In \cite{gao2018community}, the authors study the Degree-Corrected SBM, but deal with minimax performance guarantees only, and non-adaptive sampling strategies.

%% file: lower_bounds.tex
\section{Information-theoretical limits}
\label{section:info-theory-limit}

\subsection{Uniform selection strategy} 

Recall that an algorithm $\pi$ is {\it uniformly good} if for all $\set{M} \in \Omega$ and $i \in \set{I}$, $\varepsilon^\pi_i (n, T) = o(1)$ as $T \to \infty$ under $T=\omega(n)$. Assumptions (A1) and (A2) ensure the existence of uniformly good algorithms. The algorithm we present in Section \ref{section4} is uniformly good under these assumptions. The following theorem provides a lower bound on the error rate of uniformly good algorithms. 

\medskip
\begin{theorem} \label{thm:lower-random}
If an algorithm $\pi$ with uniform selection strategy is uniformly good,
then for any $\set{M}\in \Omega$ satisfying \textnormal{(A1)} and \textnormal{(A2)}, under $T=\omega(n)$, the following holds:
\begin{align*}
\liminf_{T \to \infty}  \frac{\frac{Tw}{n} 
\set{D}_{\set{M}}^U(i)}{ \log (1/\varepsilon^\pi_{i}(n, T))}
\ge  1,  \qquad \forall i \in \set{I}.%
\end{align*}
\end{theorem}
\updated{The proof of Theorem~\ref{thm:lower-random} will be presented later in this section.}
Theorem~\ref{thm:lower-random} implies that the global error rate of any uniformly good algorithm satisfies:
$$
\varepsilon^\pi(n, T) \ge 
 \exp \left( - \frac{Tw}{n} \tilde{\set{D}}^U_{\set{M}} (1+ o(1)) \right) \; ,
$$
\begin{align*}
    & \hbox{where}\quad\tilde{\set{D}}^U_{\set{M}}  :=  - \frac{n}{Tw} \log \left(
\frac{1}{n} \sum_{i \in \set{I}} \exp\left( - \frac{Tw}{n} \set{D}^U_{\set{M}}(i) \right)  \right)
\\
& \updated{\hbox{and} \quad \set{D}_{\set{M}}^U(i) := 
 \min_{k' \neq \sigma(i)}
\min_{h' \in [(h_* + 1)/2, 1]}
\frac{1}{L}\sum_{\ell} \KL(h'p_{k'\ell} + \bar{h}'\bar{p}_{k'\ell}, q_{i\ell}) > 0.}
\end{align*}
\medskip
\noindent
{\bf Divergence $\set{D}_{\set{M}}^U(i)$ and its properties.}  The {\it divergence} $\set{D}_{\set{M}}^U(i)$, defined in Section 1, quantifies the hardness of classifying item $i$. This divergence naturally appears in the change-of-measure argument used to establish Theorem \ref{thm:lower-random}. To get a better understanding of $\set{D}_{\set{M}}^U(i)$, and in particular to assess its dependence on the various system parameters, we provide the following useful upper and lower bounds, proved in \updated{Appendix~\ref{sec:pf-bound-D-M}}:

\medskip
\begin{proposition} 
\label{prop:bound-D-M}
\begin{linenomath}
Fix $i\in \set{I}$. Let $k'$ be such that:
$$
\set{D}_{\set{M}}^U(i) = \min_{h' \in [(h_* + 1)/2, 1]}\frac{1}{L}\sum_{\ell} \KL(h'p_{k'\ell} + \bar{h}'\bar{p}_{k'\ell}, q_{i\ell}).
$$
Then, we have:
\begin{align} \label{eq:bound-D-M}
& \set{D}_{\set{M}}^U(i)  \leq  \frac{1}{2 L \eta}(2h_i-1)^2
\min_{\updated{\frac{h_*}{2 h_i - 1} \le c \le \frac{1}{2 h_i -1}}}\|c\vec{r}_{k'} - \vec{r}_{\sigma(i)}\|_2^2, \nonumber
\\
&  \set{D}_{\set{M}}^U(i)  \ge 
 \frac{1}{2L} (2h_i - 1)^2 \min_{\updated{\frac{h_*}{2 h_i - 1} \le c \le \frac{1}{2 h_i -1}}} \|c\vec{r}_{k'} - \vec{r}_{\sigma(i)}\|_2^2 \;.
\end{align}
\end{linenomath}
\end{proposition}
Note that $\set{D}^U_{\set{M}}(i)$ vanishes as $h_i$ goes to $1/2$, which makes sense since for $h_i\approx 1/2$, item $i$ is very hard to cluster. We also have $\set{D}_{\set{M}}^U(i) = 0$ when $\min_{\updated{\frac{h_*}{2 h_i - 1} \le c \le \frac{1}{2 h_i -1}}}\|c\vec{r}_{k'} - \vec{r}_{\sigma(i)}\|_2^2 = 0$. In this case, $\rho_* = 0$ and there exists $h' \in [(1+h_*)/2, 1]$ such that for some $k' \neq \sigma(i)$, $2q_{i\ell} -1 = (2h'-1) (2{p}_{k'\ell} -1)$ for all $\ell \in [L]$, so that clustering item $i$ is impossible. 

\medskip
\noindent
{\bf Application to the simpler model of \cite{oh2016}.}  Consider a model with a single question and two clusters of items. From Theorem~\ref{thm:lower-random}, we can recover an asymptotic version of Theorem~$2.4.$ in \cite{oh2016}. 
\begin{corollary}
	\begin{linenomath}
	Let $L=1, K=2$, \updated{$\vec{p} = (p_{11}, p_{21})$}, and $w=1$. If an algorithm $\pi$ with uniform selection strategy is uniformly good, whenever $\set{M}$ satisfies (A1) and (A2), under $T = \omega(n)$, we have:
\begin{align*}
\liminf_{T \to \infty}  \frac{\frac{T}{n} 
	C (2 h_i - 1)^2 (\updated{p_{11} - p_{21}})^2}{ \log (1/\varepsilon^\pi_{i}(n, T))}
\ge  1, \qquad  \forall i \in \set{I},%
\end{align*}
	where $C>0$ is an absolute constant.
	\end{linenomath}
	\label{cor:lowerbound-uniform}
\end{corollary}
\updated{The proof of Corollary~\ref{cor:lowerbound-uniform} is presented in Appendix~\ref{prf:cor_simple_uniform}.}
Corollary~\ref{cor:lowerbound-uniform} implies, 
\begin{align*}
\varepsilon^\pi_{i}(n, T) \ge \exp\left( - \frac{T}{n} C (2 h_i - 1)^2 (\updated{p_{11} - p_{21}})^2 (1 + o(1))\right)
\end{align*}
as $T \to \infty$ under $T = \omega(n)$.
Smaller $h_i$ and $\updated{|p_{11} - p_{21}|}$ imply item $i$ is harder to classify. Note that Theorem~$2.4.$ in \cite{oh2016} (corresponds to $\updated{p_{21} = 1- p_{11}}$ in our \updated{Corollary~\ref{cor:lowerbound-uniform}}) provides a minimax lower bound whereas our result is problem-specific and hence more precise. 
\updated{Note that Corollary~\ref{cor:lowerbound-uniform} also applies directly to Example~\ref{ex:1} mentioned in Introduction. The lower bound on the error probability for each item $i$ scales as $\exp( -c \frac{T}{n}(2 h_i -1)^2 )$ with some constant $c>0$.}

\begin{proof}[\updated{Proof of Theorem~\ref{thm:lower-random}}]
The proof leverages change-of-measure arguments, as those used in the classical multi-armed bandit problem \cite{lai1985asymptotically} or the Stochastic Block Model \cite{yun2016optimal}. Here the proof is however complicated by the fact that we wish a lower bound on the error rate for clustering each item. 

Let $\pi$ denote a uniformly good algorithm with uniform selection strategy, and let $\set{M}\in \Omega$ be a model satisfying Assumptions (A1) and (A2). In our change-of-measure, we denote by $\set{M}$ the original model and by $\set{N}$ a perturbed model.
Fix $i \in \set{I}$, where $\sigma(i) = k$. Let $k' \in [K], h' \in[(h_*+1) /2, 1]$ denote
the minimizers for the optimization problem leading to $\set{D}_{\set{M}}^U(i)$, i.e.,
\begin{align*} 
\set{D}_{\set{M}}^U(i) =  \frac{1}{L} \sum_{\ell =1}^L \KL (h'p_{k'\ell} + \bar{h}'\bar{p}_{k'\ell}, q_{i\ell}) 
> 0
\;.
\end{align*}

For these choices of $i, k',$ and $h'$,
we construct the perturbed model $\set{N}$ as follows. Under $\set{N}$, all responses for items different than $i$ are generated as under $\set{M}$. The responses for $i$ under $\set{N}$ are generated as if $i$ was in cluster $k'$ and had difficulty $h'$. We can write the log-likelihood ratio of the observation under $\set{N}$ to that under $\set{M}$ as follows:
\begin{equation} \label{eq:likelihood-random}
\set{L} = \sum_{t=1}^T \mathbbm{1}[i \in \set{W}_t] \sum_{\ell =1}^L
 \mathbbm{1}[\ell_t = \ell]
 \left(
\mathbbm{1}[X_{i \ell t} = +1] \log \frac{q'_{\ell}}{q_{i\ell}}
+\mathbbm{1}[X_{i \ell t} = -1]  \log \frac{\bar{q}'_{\ell}}{\bar{q}_{i\ell}}
\right) \;.
\end{equation}
where we let $\vec{q}' := (q'_\ell)_{\ell \in[L]}$ with $q'_\ell = h'p_{k'\ell} + \bar{h}'\bar{p}_{k'\ell}$.

Let $\Pr_{\set{N}}$ and $\EXP_{\set{N}}$ (resp. $\Pr_{\set{M}} = \Pr$ and $\EXP_{\set{M}} = \EXP$)
denote, respectively, the probability measure and the expectation under $\set{N}$ (resp. $\set{M}$). 
Using the construction of $\set{N}$, a change-of-measure argument provides us with a connection between the error rate on item $i$ under $\set{M}$ and the mean and the variance of $\set{L}$ under $\set{N}$:
\begin{align} \label{eq:random-claim2}
\log(1/\varepsilon_i^\pi (n, T))  -  \log 4
\le \EXP_{\set{N}} [\set{L}] + \sqrt{2 \EXP_{\set{N}} \left[(\set{L} -  \EXP_{\set{N}} [\set{L}]) ^2 \right]} \;.
\end{align}

\medskip
\noindent
\updated{{\bf Proof of \eqref{eq:random-claim2}.}} The distribution of the log-likelihood $\set{L}$ under $\set{N}$ satisfies:
for any $g \ge 0 $, %
\begin{align}
\Pr_{\set{N}} \left\{\set{L} \leq g \right\}
 & = \mathbb{P}_{\set{N}} 
 \left\{\set{L} \leq g, i \in \set{E}^\pi \right\} + \mathbb{P}_{\set{N}} \left\{\set{L} \leq g, i \notin \set{E}^\pi \right\}
\nonumber \\
& \leq {\mathbb{P}_{\set{N}} \left\{\set{L} \leq g, i \in \set{E}^\pi \right\}} + {\mathbb{P}_{\set{N}} \left\{  i \notin \set{E}^\pi \right\}}.
\label{eq:likelihood-error-connect}
\end{align}
Using the definition \eqref{eq:likelihood-random} of the log-likelihood ratio, we bound the first term in \eqref{eq:likelihood-error-connect} as follows:
\begin{align}
{\mathbb{P}_{\set{N}} \left\{\set{L} \leq g, i \in \set{E}^\pi \right\}} & = \int_{\left\{ \set{L} \leq g, i \in \set{E}^\pi \right\}} d\mathbb{P}_{\set{N}} \nonumber \\ 
& = \int_{\left\{ \set{L} \leq g, i \in \set{E}^\pi \right\}} \exp(\set{L}) d\mathbb{P}_{\set{M}} \nonumber \\ 
& \leq \exp(g) \mathbb{P}_{\set{M}} \left\{\set{L} \leq g, i \in \set{E}^\pi \right\} \nonumber\\ 
& \leq \exp(g) \varepsilon^\pi_i(n, T). \label{eq:random-a-term}
\end{align}

To bound the second term in \eqref{eq:likelihood-error-connect},
note that $(2h' -1)$ is a strictly positive constant.\footnote{This is true as $h'$ is optimized over $[(h_*+1)/2, 1]$. }
Hence, the perturbed model ${\set{N}}$ satisfies (A1). By the definition of the uniformly good algorithm, we have ${\mathbb{P}_{\set{N}} \left\{ i \notin \set{S}_{k'}^\pi \right\}}  = o(1)$. Hence: 
\begin{align} 
{\mathbb{P}_{\set{N}} \left\{  i \notin \set{E}^\pi \right\}} = {\mathbb{P}_{\set{N}} \left\{ i \in {\set{S}}_{k}^\pi \right\}} \leq  {\mathbb{P}_{\set{N}} \left\{ i \notin {\set{S}}_{k'}^\pi \right\}} \leq \frac{1}{4}.
\label{eq:random-b-term}
\end{align}
Combining 
\eqref{eq:likelihood-error-connect}, \eqref{eq:random-a-term} and \eqref{eq:random-b-term} with $g = - \log(4 \varepsilon_i^\pi (n, T))$,
we have
\begin{align}
\mathbb{P}_{\set{N}} \left\{\set{L} \leq 
- \log \varepsilon^\pi_i (n,T) - \log 4
 \right\} & \leq \frac{1}{2}. \label{eq:pluggedloglikelihodd}
\end{align}
Using Chebyshev's inequality, we obtain:
\begin{align*}
\mathbb{P}_{\set{N}} \left\{
\set{L} - \mathbb{E}_{\set{N}}[\set{L}]   \geq  \sqrt{2 \mathbb{E}_{{\set{N}}} [(\set{L} - \mathbb{E}_{\set{N}}[\set{L}] )^2 ]} \right\}
\leq 
\frac{\mathbb{E}_{{\set{N}}} [(\set{L} - \mathbb{E}_{\set{N}}[\set{L}] )^2 ]}{
\left(
\sqrt{2 \mathbb{E}_{{\set{N}}} [(\set{L} - \mathbb{E}_{\set{N}}[\set{L}] )^2 ]}  \right)^2
}
= \frac{1}{2} \;,
\end{align*}
which implies $\mathbb{P}_{\set{N}} \left\{
\set{L}    \leq \mathbb{E}_{\set{N}}[\set{L}]+ \sqrt{2 \mathbb{E}_{{\set{N}}} [(\set{L} - \mathbb{E}_{\set{N}}[\set{L}] )^2 ]} \right\}
\ge  \frac{1}{2}$.
Combining this result to \eqref{eq:pluggedloglikelihodd}
implies the claim \eqref{eq:random-claim2}. 

\medskip
\noindent
\updated{Next, }Lemma~\ref{lem:random-log-calc}  
provides the upper bound on mean and variance of $\set{L}$ under \updated{the model} ${\set{N}}$.
\begin{lemma}\label{lem:random-log-calc}
Assume that \textnormal{(A2)} holds. For $i, i'$ such that $\sigma(i) = k \neq k' = \sigma(i')$,
under the uniform selection strategy, we have
\begin{align*} 
\EXP_{\set{N}}[\set{L}] &=  \frac{Tw}{n} \set{D}_{\set{M}}^U (i)
\;, 
\qand \\
\EXP_{\set{N}} \left[(\set{L} -  \EXP_{\set{N}} [\set{L}]) ^2 \right]
&\le 
\frac{Tw}{Ln} 
\log\left(\frac{1}{\eta}\right) \sum_{\ell = 1}^L \KL(q'_{\ell}, q_{i \ell}) 
+ \sqrt{\KL(q_{i'\ell}, q_{i \ell})}
 \;.
\end{align*}
\end{lemma}
The proof of this lemma is presented to Appendix~\ref{sec:pf-random-log-calc}.
Note that in view the above lemma, the r.h.s. of \eqref{eq:random-claim2} is asymptotically dominated by 
$\EXP_{\set{N}}[\set{L}]$, since  $\EXP_{\set{N}}[\set{L}] = \Omega(T/n)$ and $\sqrt{ \EXP_{\set{N}} \left[(\set{L} -  \EXP_{\set{N}} [\set{L}]) ^2 \right]} = O(\sqrt{T/n})$.
Thus, Theorem~\ref{thm:lower-random} follows from the claim in \eqref{eq:random-claim2} and Lemma~\ref{lem:random-log-calc}. \ep

\end{proof}

\subsection{Adaptive selection strategy} 

The derivation of a lower bound for the error rate under adaptive (list, pair) selection strategies is similar:

\medskip
\begin{theorem} \label{thm:lower-adaptive} 
For any $\set{M}\in \Omega$ satisfying \textnormal{(A1)} and \textnormal{(A2)}, and for any uniformly good algorithm $\pi$ with possibly adaptive (list, question) selection strategy, under $T=\omega(n)$, we have: 
\begin{linenomath}
\begin{align*}
\liminf_{T \to \infty} 
\frac{\frac{Tw}{n} 
\tilde{\set{D}}_{\set{M}}^A}{ \log (1/\varepsilon^\pi(n, T))}
\ge  1,
\end{align*}
\updated{
\begin{align*}
&\hbox{where } \ \ \ \tilde{\set{D}}_{\set{M}}^A := 
\max_{\vec{y} \in \set{Y}(n)} - \frac{n}{Tw}\log \left( \frac{1}{n} \sum_{i=1}^n \exp \left( - \frac{Tw}{n} {\set{D}}_{\set{M}}^A (i, \vec{y}) \right) \right),
\\
&\ \ \ \ \ \ \ \ \ \ \ \ \ \  {\set{D}}_{\set{M}}^A (i, \vec{y}) :=
\min_{ j : \sigma(j) \neq \sigma(i)} \sum_{\ell} \left( y_{j\ell} \KL( q_{j \ell}, q_{i\ell})+y_{i\ell} \KL( q_{i \ell}, q_{j\ell}) \right),\\
&\hbox{and } \ \ \ \set{Y}(n) :=  \left\{
\vec{y} \in [0, 1]^{n\times L} :  \sum_{i \in \set{I}, \ell \in [L]} y_{i \ell} = n
\right\}.
\end{align*}}
\end{linenomath}
\end{theorem}
Theorem~\ref{thm:lower-adaptive} implies $	\varepsilon^\pi(n, T) \ge 
\exp ( - \frac{Tw}{n} \tilde{\set{D}}_{\set{M}}^A (1+ o(1)) ). $

\begin{proof}[\updated{Proof of Theorem~\ref{thm:lower-adaptive}}]

Again we use a change-of-measure argument, where we swap two items from different clusters. First, we prove the lower bound for the error rate of a fixed item $i$. Fix $i\in \set{I},$ let $j$ be an item satisfying $\sigma(j) \neq \sigma(i)$ and let
\begin{align*}
\set{D}^A_{\set{M}}(i, \vec{y}) =  \sum_{\ell =1}^L
\left({y}_{j \ell} \KL (q_{j\ell}, q_{i \ell}) +{y}_{i \ell} \KL (q_{i\ell}, q_{j \ell})  \right).
\end{align*}
$\set{D}^A_{\set{M}}(i, \vec{y}) $ is the value of the optimization problem: 
$$
\min_{ j : \sigma(j) \neq \sigma(i)} \sum_{\ell} \left( y_{j\ell} \KL( q_{j \ell}, q_{i\ell})+y_{i\ell} \KL( q_{i \ell}, q_{j\ell}) \right).
$$

\medskip
Consider a perturbed model ${\set{N}}'$, in which items except $i$ and $j$ have the same response statistics as under ${\set{M}}$, and in which item $i$ behaves as item $j$, and item $j$ behaves as item $i$. Let $\Pr_{\set{N}'}$ and $\EXP_{\set{N}' }$ denote, respectively, the probability measure and the expectation under ${\set{N}}'$. The log-likelihood ratio of the responses under ${\set{N}}' $ and under ${\set{M}}$ is:
\begin{align}
	\set{L} = & \sum_{t=1}^T \mathbbm{1}[i \in {\set{W}}_t] \sum_{\ell =1}^L
	\mathbbm{1}[\ell_t = \ell]
	\left(
	\mathbbm{1}[X_{i \ell t} = +1] \log \frac{q_{j\ell}}{q_{i \ell}}
	+\mathbbm{1}[X_{i \ell t} = -1]  \log \frac{\bar{q}_{j\ell}}{\bar{q}_{i \ell}}
	\right) + \nonumber
	\\
	&\sum_{t=1}^T \mathbbm{1}[j \in {\set{W}}_t] \sum_{\ell =1}^L
	\mathbbm{1}[\ell_t = \ell]
	\left(
	\mathbbm{1}[X_{j \ell t} = +1] \log \frac{q_{i\ell}}{q_{j \ell}}
	+\mathbbm{1}[X_{j \ell t} = -1]  \log \frac{\bar{q}_{i\ell}}{\bar{q}_{j \ell}}
	\right) \;.
\end{align}
The mean and variance of $\set{L}$ under ${\set{N}}'$ are:
\begin{align*} 
\EXP_{{\set{N}}' }[\set{L}]  = &\sum_{\ell =1}^{L}\left( \EXP_{{\set{N}}' }[Y_{i \ell}] \KL(q_{j\ell}, q_{i \ell})+\EXP_{{\set{N}}' }[Y_{j \ell}] \KL(q_{i\ell}, q_{j \ell}) \right)
\\
 = & \frac{Tw}{n} \sum_{\ell =1}^L \left( y_{j \ell} \KL(q_{j\ell}, q_{i \ell})+y_{i \ell} \KL(q_{i\ell}, q_{j \ell}) \right)
\\
= & \frac{Tw}{n} \set{D}_{\set{M}}^A ( i, \vec{y}), 
\end{align*}
\begin{align*}
\EXP_{{\set{N}}' } \left[(\set{L} -  \EXP_{{\set{N}}} [\set{L}]) ^2 \right]
 \le &  \log\left(\frac{1}{\eta}\right) \sum_{\ell =1}^{L} \EXP_{{\set{N}}'}[Y_{i \ell}] \left( \KL(q_{j\ell}, q_{i \ell}) 
+\sqrt{\KL(q_{j\ell}, q_{i \ell})} \right ) + 
\\
& \log\left(\frac{1}{\eta}\right) \sum_{\ell =1}^{L} \EXP_{{\set{N}}'}[Y_{j \ell}] \left( \KL(q_{i\ell}, q_{j \ell}) 
+\sqrt{\KL(q_{i\ell}, q_{j \ell})} \right )
\\
 = & \frac{Tw}{n} 
\log\left(\frac{1}{\eta}\right) \sum_{\ell = 1}^L 
y_{j \ell} \left(
\KL(q_{j\ell}, q_{i \ell}) 
+\sqrt{\KL(q_{j\ell}, q_{i \ell})} \right)+\\
& \frac{Tw}{n} 
\log\left(\frac{1}{\eta}\right) \sum_{\ell = 1}^L 
y_{i \ell} \left(
\KL(q_{i \ell}, q_{j \ell}) 
+\sqrt{\KL(q_{i \ell}, q_{j \ell})} \right),
\end{align*}
using a slight modification of Lemma \ref{lem:random-log-calc}. By a similar argument as that used in the proof of Theorem \ref{thm:lower-random}, we get:
\begin{align} \label{eq:adaptive-claim2}
\log(1/\varepsilon_{i}^\pi (n, T))  -  \log 4
\le \EXP_{\set{N}' } [\set{L}] + \sqrt{2 \EXP_{\set{N}' } \left[(\set{L} -  \EXP_{\set{N}' } [\set{L}]) ^2 \right]} \;,
\end{align}
as is in \eqref{eq:random-claim2}. Note that the r.h.s. of \eqref{eq:adaptive-claim2} is asymptotically dominated by 
$\EXP_{\set{N}' }[\set{L}]$ as $\EXP_{\set{N}' }[\set{L}] = \Omega(T/n)$ and $\sqrt{ \EXP_{\set{N}'} \left[(\set{L} -  \EXP_{\set{N}' } [\set{L}]) ^2 \right]} = O(\sqrt{T/n})$. 
That is, 
\begin{align*}
	\EXP_{\set{N}' } [\set{L}] + \sqrt{2 \EXP_{\set{N}' } \left[(\set{L} -  \EXP_{\set{N}' } [\set{L}]) ^2 \right]} & =  \frac{Tw}{n}{\set{D}}_{\set{M}}^A(i, \vec{y})  (1+ o(1)) \;.
\end{align*}
We deduce that $\varepsilon_{i}^\pi (n, T) \ge \exp\left(  -  \frac{Tw}{n}{\set{D}}_{\set{M}}^A(i, \vec{y})  (1+ o(1)) \right)$. Thus, from the definition of $\tilde{D}_{\set{M}}^A$, we have: 
\begin{align*}
	\varepsilon^\pi (n, T) 	& = \frac{1}{n}\sum_{i \in \set{I}} \varepsilon_{i}^\pi (n, T)
	\\
	& = \frac{1}{n} \sum_{i \in \set{I}} \exp \left( -\frac{Tw}{n}  {\set{D}}_{\set{M}}^A (i, \vec{{y}})  (1 + o(1)) \right)
	\\
	& \ge 	\exp \left( - \frac{Tw}{n} {D}_{\set{M}}^A(1+ o(1)) \right),
\end{align*}
Taking the logarithm of the previous inequality, we conclude the proof. \ep
\end{proof}

%% file: algorithms.tex
\section{Algorithms}\label{section4}

In this section, we describe our algorithms for both uniform and adaptive (list, question) selection strategies.

\begin{algorithm}
	\begin{algorithmic}
		\STATE {\bfseries Input:} $K, T$
		\medskip
		\STATE {\bfseries User response collection:}
		\STATE Question $\ell$ is asked to each item $i$ for $\tau = \lfloor \frac{T w}{n L} \rfloor$ times.
		\STATE For all $\ell \in [L]$ and $i \in \set{I}$, $x_{i \ell } \gets \sum_{t}^T \mathbbm{1}[i \in \set{W}_t, \ell_t = \ell, X_{i \ell t} = 1]$  and $\hat{q}_{i \ell } \gets x_{i \ell } / \tau$.
		\medskip
		\STATE {\bfseries Normalization: } 
		\STATE $\hat{\vec r}^{i} \gets \frac{2 \hat{\vec q}_{i} - 1}{\|2 \hat{\vec q}_{i} - 1\|}$ for all $i \in \set{I}$
		\medskip
		\STATE {\bfseries K-means clustering: }  
		\STATE $T_{i} \gets \{ j \in \set{I} : \|\hat{\vec r}^{j} - \hat{\vec r}^{i} \|^2 \leq \left(\frac{n}{T}\right)^{\frac{1}{2}}\}$ for each $i \in \set{I}$
		\STATE $\set{S}_0  \gets \emptyset$
		\FOR{$k=1$ {\bfseries to} $K$}
		\STATE $i_k^\star \gets \argmax_{i \in \set{I}} |T_i \setminus \bigcup_{k' = 0}^{k -1} \set{S}_{k'} | $
		\STATE $\set{S}_k \gets T_{i_k^\star} \setminus \bigcup_{k' = 0}^{k-1} \set{S}_{k'}$
		\STATE $\vec{\xi}_k \gets  (\sum_{i \in \set{S}_k } \hat{\vec r}^i ) /  |\set{S}_k| $
		\ENDFOR
		\STATE For each $i \in \set{I} \setminus \bigcup_{k=1}^K \set{S}_k$ (or each $i\in \set{I}$),
		place $i$ in $\set{S}_{k^\star}$, where $k^\star = \argmin_{k} \|\vec{\xi}_k  - \hat{\vec r}^i \|$
		\medskip
		\STATE {\bfseries Output:}  $\{ \set{S}_k \}_{k \in [K]} $
	\end{algorithmic}
	\caption{Uniform (list, question) selection. }
	\label{alg:kmeans_for_r}
\end{algorithm}

\subsection{Uniform selection strategy}
In this case, we assume that with a budget of $T$ users, each item receives the same amount of answers for each question. After gathering these answers, we have to exploit the data to estimate the clusters. To this aim, we propose an extension of the K-means clustering algorithm, that efficiently leverages the problem structure. The pseudo-code of the algorithm is presented in Algorithm~\ref{alg:kmeans_for_r}. 

The algorithm first estimates the parameters ${q_{\ell i}}$: the estimator $\hat{q}_{\ell i}$ just counts the number of times item $i$ has received a positive answer for question $\ell$. We denote by $\hat{\vec{q}}_i=(\hat{q}_{\ell i})_\ell$ the resulting vector. By normalizing the vector $2 \hat{\vec{q}}_i -1$, we can decouple the nonlinear relationship between $\vec{q}$, $\vec h$ and $\vec p$. Let 
$\hat{\vec r}^i = \frac{2 \hat{\vec q}_{i} - 1}{\|2 \hat{\vec q}_{i} - 1\|}$ be the normalized vector. Then,
$\hat{\vec r}^i$ concentrates around $\tilde{\vec r}_{\sigma(i)}:=\vec{r}_{\sigma(i)}/\|\vec{r}_{\sigma(i)}\|$. Importantly, the normalized vector $\tilde{\vec r}_{\sigma(i)}$ does not depend on $h_i$ but on the cluster index $\sigma (i)$ only. The algorithm exploits this observation, and applies the K-means algorithm to cluster the vectors $\hat{\vec r}^i$. By analyzing how $\hat{\vec r}^i$ concentrates around $\tilde{\vec r}_{\sigma(i)}$ and by applying the results to our properly tuned algorithm (decision thresholds), we establish the following theorem.
\begin{theorem}
	Assume $T= \omega\left( n\right)$ and $T = o (n^{2})$. Under Algorithm \ref{alg:kmeans_for_r}, we have,
	\begin{linenomath}
	\begin{align}
\varepsilon_i(n,T)\leq \exp \left( - \frac{(2 h_i -1)^2  \rho_*^2  }{200} \frac{Tw}{Ln}(1 + o(1))\right).
\end{align} 
\end{linenomath}
	\label{thm:misclassification_after_phase_1}
\end{theorem}
\updated{We will present the proof of Theorem~\ref{thm:misclassification_after_phase_1} later in this section.}
In view of Proposition~\ref{prop:bound-D-M} and the lower bounds derived in the previous section, we observe that the exponent for the mis-classification error of item $i$ has the correct dependence in $Tw/Ln$ and the tightest possible scaling in the hardness of the item, namely $(2h_i-1)^2$. Also note that using Proposition~\ref{prop:bound-D-M}, the equivalence between the $\ell_\infty$-norm and the Euclidean norm, and (A1), we have: $\set{D}_{\set{M}}^U(i) \geq C \frac{(2 h_i - 1)^2}{L} \rho_*^2$, for some absolute constant $C>0$. Hence, Algorithm~\ref{alg:kmeans_for_r} has a performance scaling optimally  w.r.t. all the model parameters.

The computational complexity of Algorithm~\ref{alg:kmeans_for_r} is $\set{O}(n^2)$. By choosing a small ($\log n$) subset of items (and not all the items in $\set{I}$) to compute centroids ($T_i$), it is possible to reduce the computational complexity to $\set{O}(n \log n)$. This would not affect the performance of the algorithm in practice, but would result in worse performance guarantees.

\begin{proof}[\updated{Proof of Theorem \ref{thm:misclassification_after_phase_1}}]

In this proof, we let $\tau = \lfloor \frac{T w}{n L} \rfloor$ be the number of times question $\ell$ is asked for item $i$. We also denote by $\vec{\alpha} := (\alpha_1, \ldots, \alpha_K)$ the fractions of items that are in the various clusters, i.e., $|\set{I}_k| = \alpha_k n$. Without loss of generality, and to simplify the notation, we assume that the set of misclassified items is $\mathcal{E} = \cup_{k=1}^K (\set{I}_k \setminus \set{S}_k)$, where recall that $\{ \set{S}_k\}_{k\in [K]}$ is the output of the algorithm (i.e., the permutation $\gamma$ in the definition of this set is just the identity).   

The proof proceeds in three steps: (i) we first decompose the probability of clustering error for item $i$, using the design of the algorithm and Assumptions (A1) and (A2). We show that this probability can be upper bounded by the probabilities of events related to $ \|\hat{\vec r}^i - \tilde{\vec r}_{\sigma(i)}\|$ and $\|\xi_{k} -  \tilde{\vec{r}}_{k}\|$ for all $k$, where recall that $\tilde{\vec r}_{k}:=\vec{r}_{k}/\|\vec{r}_{k}\|$. The remaining steps of the proof aim at bounding the probabilities of these events. Towards this objective, (ii) in the second step, we establish a concentration result on $ \|\hat{\vec r}^i - \tilde{\vec r}_{\sigma(i)}\|$, and (iii) the last step upper bound $\|\xi_{k} -  \tilde{\vec{r}}_{k}\|$.

\medskip
\noindent
{\it Step 1. Error probability decomposition.} The algorithm classifies item $i$ to the cluster $k$ minimizing the distance between $\hat{r}^i$ and $\xi_k$. As a consequence, we have:
\begin{align}
	 \mathbb{P}\left\{i \in \set{E} \right\}   & =  \mathbb{P}\left\{  \| \hat{\vec r}^i  - \xi_{\sigma(i)} \| \geq \min_{ k' \neq \sigma(i)}\| \hat{\vec r}^i  - \xi_{k'} \|  \right\} \nonumber
	\\
	 &\le  \mathbb{P}\left\{  \| \hat{\vec r}^i  -  \tilde{\vec r}_{\sigma(i)} \|  + \|\tilde{\vec r}_{\sigma(i)} - \xi_{\sigma(i)} \| \geq \min_{ k' \neq \sigma(i)}\left\{ \| \hat{\vec r}^i  -  \tilde{\vec r}_{k'} \| -\|  \tilde{\vec r}_{k'} - \xi_{k'} \|  \right\} \right\} \nonumber
	\\
	&\le  \mathbb{P}\left\{  \exists k' \neq \sigma(i)  : \ \ \   \|\tilde{\vec r}_{\sigma(i)} - \xi_{\sigma(i)} \|+\|  \tilde{\vec r}_{k'} - \xi_{k'} \| +2 \| \hat{\vec r}^i  -  \tilde{\vec r}_{\sigma(i)} \| \geq \| \tilde{\vec r}_{\sigma(i)} -\tilde{\vec r}_{k'} \|  \right\}, \nonumber
\end{align}	
where the two above inequalities are obtained by simply applying the triangle inequality. Now observe that in view of Assumptions (A1) and (A2), we have: for $k' \neq \sigma(i)$,
\begin{align*}
\| \tilde{\vec r}_{\sigma(i)} - \tilde{\vec r}_{k'}\| & = \frac{1}{\| {\vec r}_{\sigma(i)}\|} \left\| \vec{r}_{\sigma(i)} - \frac{\|\vec{r}_{\sigma(i)} \|}{\|\vec{r}_{k'} \|} \vec{r}_{k'} \right\| \geq \frac{\rho_*}{\| {\vec r}_{\sigma(i)}\|}= \frac{\rho_*}{ \|2 \vec{p}_{\sigma(i)} -1\|} \geq  \frac{\rho_*}{|2 \eta -1|}.
\end{align*}
We deduce that:
\begin{align}
\mathbb{P}\left\{i \in \set{E} \right\}   & \le\mathbb{P}\left\{  \exists k' \neq \sigma(i)  : \ \ \   \|\tilde{\vec r}_{\sigma(i)} - \xi_{\sigma(i)} \|+\|  \tilde{\vec r}_{k'} - \xi_{k'} \| +2 \| \hat{\vec r}^i  -  \tilde{\vec r}_{\sigma(i)} \| \geq  \frac{\rho_*}{ \|2 \vec{p}_{\sigma(i)} -1\|} \right\}, \nonumber \\
& \le  \mathbb{P}\left\{  \| \hat{\vec r}^i  - \tilde{\vec r}_{\sigma(i)} \| \geq \frac{\rho_{*}}{4 \|2 \vec{p}_{\sigma(i)} -1\|} \textrm{ or } \max_{1\leq k' \leq K} \| \xi_{k'} - \tilde{\vec r}_{k'} \| \geq \frac{\rho_{*}}{4 \|2 \vec{p}_{\sigma(i)} -1\|} \right\} \nonumber
	\\
	& \le \underbrace{\mathbb{P}\left\{ \| \hat{\vec r}^i  - \tilde{\vec{r}}_{\sigma(i)} \| \geq \frac{\rho_{*}}{4 \|2 \vec{p}_{\sigma(i)} -1\|} \right\} }_{\hbox{term }(a)} +\underbrace{ \sum_{k' =1}^{K}\mathbb{P}\left\{  \|\xi_{k'} -  \tilde{\vec{r}}_{k'}\| \geq \frac{\rho_{*}}{4 |2 \eta -1|} \right\} }_{\hbox{term }(b)}.\nonumber
\end{align}
	
\medskip
\noindent 
{\it Step 2. Concentration of $\hat{r}^i$ and upper bound on $(a)$.} We prove the following lemma, a consequence of a concentration result of $\hat{\vec{q}}_i$:  %
\begin{lemma}
	Let $0 < \varepsilon \le \frac{\|2 \vec{q}_i - 1\|}{16}$, $\tilde{\vec r}^i = \frac{{\vec r}_{\sigma(i)}}{\|{\vec r}_{\sigma(i)}\|}$, and $\tilde{\vec r}_k = \frac{{\vec r}_{k}}{\|{\vec r}_{k}\|}$. For each $i \in \set{I}$,
	\begin{align*}
	\|\hat{\vec r}^i - \tilde{\vec r}^i\| \leq \frac{5 \varepsilon }{\|2 {\vec q}_i -1  \|} ,
	\end{align*}
	with probability at least  $1 - 2 L \exp \left( -{2 \tau } \varepsilon^2 \right)$.
	\label{lm:concentration_of_r}
\end{lemma}
\updated{The proof of Lemma~\ref{lm:concentration_of_r} is presented at Appendix~\ref{sec:prooflems}.}
Note that by definition of $\rho_*$, we have:
\begin{align*}
{(2 h_i -1)   \rho_{*}} \leq {(2 h_i -1)   \|\vec{r}_{\sigma(i)}\|} = {\|2 \vec{q}_i - 1\|}.
\end{align*}
Applying Lemma \ref{lm:concentration_of_r} with $\varepsilon = \frac{(2 h_i -1)   \rho_{*}}{20} < \frac{\|2 \vec{q}_i -1\|}{16}$, we obtain an upper bound on the term $(a)$:
\begin{align*}
 \mathbb{P} \left\{  \| \hat{\vec r}^i  - \tilde{\vec r}_{\sigma(i)} \| \geq  \frac{\rho_{*}}{4 \|2 \vec{p}_{\sigma(i)} -1\|} \right\}
 & = \mathbb{P} \left\{  \| \hat{\vec r}^i  - \tilde{\vec r}_{\sigma(i)} \| \geq  \frac{5}{(2 h_i -1)\|(2 \vec{p}_{\sigma(i)} -1)\| } \frac{(2 h_i -1)  \rho_* }{20} \right\}
\\
& = \mathbb{P} \left\{  \| \hat{\vec r}^i  - \tilde{\vec r}_{\sigma(i)} \| \geq  \frac{5}{\|2 \vec{q}_i - 1\| } \frac{(2 h_i -1)  \rho_* }{20} \right\}
\\
& \leq 2 L \exp \left( - \frac{(2 h_i -1)^2  \rho_*^2  }{200} \frac{Tw}{Ln}\right)
\\
& = \exp \left( - \frac{(2 h_i -1)^2  \rho_*^2  }{200} \frac{Tw}{Ln}(1 + o(1))\right).
\end{align*}

\medskip
\noindent
{\it Step 3. Upper bound of the term $(b)$.} Next, we establish the following claim: 
\begin{equation}
\sum_{k' =1}^{K}\mathbb{P}\left\{  \|\xi_{k'} -  \tilde{\vec{r}}_{k'}\| \geq \frac{\rho_{*}}{4 |2 \eta -1|} \right\} \le  \exp\left( - n \left( \Theta\left(\frac{(T/n)^{\frac{1}{2}}}{\log\left( T/n\right)}\right) \right)\right).
\label{eq:calEupper}
\end{equation}
To this aim, we first show that a large fraction of the items satisfy $\|\hat{\vec r}^v -\tilde{\vec r}^v\| \leq \frac{1}{4} \left(\frac{n}{T}\right)^{\frac{1}{4}}$. Applying Lemma~\ref{lm:concentration_of_r} with $\varepsilon = \frac{\|2 \vec{q}_i - 1 \|}{20} \left(\frac{n}{T}\right)^{\frac{1}{4}}$, we get:
\begin{align}
	\mathbb{P}\left\{\|\hat{\vec r}^i - \tilde{\vec r}^i \| \geq \frac{1}{4} \left( \frac{n}{T}\right)^{\frac{1}{4}} \right\} & = \Pr\left\{\|\hat{\vec r}^i - \tilde{\vec r}^i \| \geq \frac{5}{\|2 \vec{q}_i -1\|} \frac{\|2 \vec{q}_i - 1 \|}{20} \left( \frac{n}{T}\right)^{\frac{1}{4}} \right\}
	\nonumber
	\\
	& \leq 2 L \exp \left( - \frac{\|2 \vec{q}_i - 1\|^2 }{200} \frac{w}{L} \left(\frac{T}{n}\right)^{\frac{1}{2}}\right)
	\nonumber
	\\
	& \leq 2 L \exp\left(- \frac{h_*^2 \rho_*^2 w}{200 L} \left(\frac{T}{n}\right)^{\frac{1}{2}} \right). \label{eq:rhatrtil}
\end{align}
Define $p_{\max} :=  \max_{v \in \set{I}_k} \mathbb{P} \left\{ \| \hat{\vec r}^v- \tilde{\vec{r}}^v\|\geq \frac{1}{4} \left(\frac{n}{T}\right)^{\frac{1}{4}} \right\} $. Then from  \eqref{eq:rhatrtil}, $p_{\max} \leq \exp\left( -\Theta\left( \left(\frac{T}{n}\right)^{\frac{1}{2}} \right) \right)$. Further define $S$ as the number of the items in $\set{I}$ that satisfy $\|\hat{\vec r}^v -\tilde{\vec r}^v\| \leq \frac{1}{4} \left(\frac{n}{T}\right)^{\frac{1}{4}}$, i.e., 
$S = \sum_{v \in \set{I}} \mathbbm{1}_{\{\| \hat{\vec r}^v- \tilde{\vec{r}}^v\|   \geq \frac{1}{4} (\frac{n}{T})^{\frac{1}{4}} \}}$. Since $\hat{\vec r}^1, \ldots, \hat{\vec r}^n$ are independent random variables, $\mathbbm{1}_{\{\| \hat{\vec r}^1- \tilde{\vec{r}}^1\|   \geq \frac{1}{4} (\frac{n}{T})^{\frac{1}{4}} \}}, \ldots, \mathbbm{1}_{\{\| \hat{\vec r}^n- \tilde{\vec{r}}^n\| \geq \frac{1}{4} (\frac{n}{T})^{\frac{1}{4}} \}}$ are independent Bernoulli random variables. 
From Chernoff bound, we get:
\begin{align*}
\mathbb{P} \left\{ S \geq  \frac{n}{\log \left( \frac{T}{n}\right)} \right\} & \leq \inf_{\lambda >0 } \frac{\mathbb{E}[\exp(\lambda S )]}{\exp\left( \lambda \frac{n}{\log \left( \frac{T}{n}\right)} \right)} \leq \inf_{\lambda >0 } \frac{(1 +p_{\max} (e^\lambda -1))^{ n}}{\exp\left( \lambda \frac{n}{\log \left( \frac{T}{n}\right)} \right)}
\\
& \leq \inf_{\lambda >0 } \exp \left(  n p_{\max}  (e^\lambda -1) - \lambda \frac{n}{\log \left( \frac{T}{n}\right)}\right)
\\
& \stackrel{(i)}{\leq} \exp\left(  n p_{\max} \left( \frac{1}{p_{\max}} - 1\right) -\log \frac{1}{p_{\max}} \frac{n}{\log \left( \frac{T}{n}\right)} \right) 
\\
& = \exp\left( - n \left( \Theta\left(\frac{(T/n)^{\frac{1}{2}}}{\log\left( T/n\right)}\right) +  p_{\max} - 1 \right)\right)
\\
& \leq \exp\left( - n \left( \Theta\left(\frac{(T/n)^{\frac{1}{2}}}{\log\left( T/n\right)}\right) \right)\right)
\end{align*}
where for $(i)$, we set $\lambda = \log \frac{1}{p_{\max}}$. Therefore, to prove (\ref{eq:calEupper}), it suffices to show that: 
\begin{align*}
\max_{1\le k \le K}\|\xi_{k} -  \tilde{\vec{r}}_{k}\| \leq \frac{\rho_{*}}{4 |2 \eta -1|}  \quad \mbox{when}\quad S \leq  \frac{n}{\log \left( \frac{T}{n}\right)} .
\end{align*}

Assume that $S \leq  \frac{n}{\log \left( \frac{T}{n}\right)} $. Then, every $v$ having $\min_{1\le k\le K}\|\hat{\vec r}^v - \tilde{\vec r}_k \| \geq 2\left( \frac{n}{T}\right)^{\frac{1}{4}}$ cannot be a center node (i.e., one the $i_k^*$ for $k=1,\ldots,K$). This is due to the following facts:
\begin{itemize}
\item[(i)] $|T_v| \leq \frac{n}{\log \left( \frac{T}{n}\right)}$ when $\min_{1\le k\le K}\|\hat{\vec r}^v - \tilde{\vec r}_k \| \geq 2 \left( \frac{n}{T}\right)^{\frac{1}{4}}$, since for all $w$ such that $\| \hat{\vec r}^w- \tilde{\vec{r}}^w\|   \leq \frac{1}{4} (\frac{n}{T})^{\frac{1}{4}}$,
$
\|\hat{\vec r}^v - \hat{\vec r}^w \| \geq \|\hat{\vec r}^v - \tilde{\vec r}^w \|- \|\tilde{\vec r}^w - \hat{\vec r}^w \| \geq \frac{3}{2} \left( \frac{n}{T}\right)^{\frac{1}{4}}.
$
\item[(ii)] $|T_v| \geq \alpha_k n - \frac{n}{\log \left( \frac{T}{n}\right)}$ when $\|\hat{\vec r}^v - \tilde{\vec r}_k \| \leq \frac{1}{2} \left( \frac{n}{T}\right)^{\frac{1}{4}}$, since for all $w\in \set{I}_k$ such that $\| \hat{\vec r}^w- \tilde{\vec{r}}^w\|   \leq \frac{1}{4} (\frac{n}{T})^{\frac{1}{4}}$,
$
\|\hat{\vec r}^v - \hat{\vec r}^w \| \leq \|\hat{\vec r}^v - \tilde{\vec r}_k \| + \|\tilde{\vec r}_k - \hat{\vec r}^w \| \leq \frac{3}{4} \left( \frac{n}{T}\right)^{\frac{1}{4}}.
$
\end{itemize}
Therefore, when $\frac{T}{n} = \omega(1)$,
\begin{align}
\|\hat{\vec r}_{i^\star_k} - \tilde{\vec r}_k \|  \leq 2\left( \frac{n}{T}\right)^{\frac{1}{4}}. \label{eq:rstarup}
\end{align}

Let $\set{R}_k$ denote the set of items $\set{S}_k$ before computing $\xi_k$ ($\set{S}_k$ used for the calculation of $\xi_k$) -- see the algorithm. Then, from \eqref{eq:rstarup} and the definition of $\set{S}_k$ before computing $\xi_k$, 
\begin{align*}
\|\hat{\vec r}^v - \tilde{\vec r}_k \| \leq \|\hat{\vec r}^v - \hat{\vec r}_{i^\star_k} \| +\|\hat{\vec r}_{i^\star_k} - \tilde{\vec r}_k \|  \leq 3 \left( \frac{n}{T}\right)^{\frac{1}{4}} \quad \mbox{for all} \quad v \in \set{R}_k.
\end{align*}
From the above inequality and Jensen's inequality, 
\begin{align*}
	\| \xi_k - \tilde{\vec{r}}_{k}\| &  = \left\| \frac{\sum_{v \in \set{R}_k} \hat{\vec r}^v}{|\set{R}_k|} - \tilde{\vec{r}}_{k}\right\|
	\\
	& \leq \frac{1}{| \set{R}_k|} \left( \sum_{v \in \set{R}_k } \| \hat{\vec r}^v - \tilde{\vec{r}}_{k} \| \right)
	\leq 3 \left(\frac{n}{T}\right)^{\frac{1}{4}} .
\end{align*}
Therefore, when $T = \omega (n)$,
\begin{align*}
\max_{1\le k \le K}\|\xi_{k} -  \tilde{\vec{r}}_{k}\| \leq \frac{\rho_{*}}{4  | 2 \eta -1|}  \quad \mbox{when}\quad S \leq  \frac{n}{\log \left( \frac{T}{n}\right)} ,
\end{align*}
which concludes the proof of (\ref{eq:calEupper}).

\medskip
\noindent
The proof of theorem is completed by remarking that when $T=o(n^2)$, then 
$$
\frac{T}{n} = o\left( n \frac{(T/n)^{\frac{1}{2}}}{\log\left( T/n\right)}\right).
$$
This implies that the upper bound we derived for the term $(a)$ is dominating the upper bound of the term $(b)$. Finally,
$$
\mathbb{P}\left\{i \in \set{E} \right\} \le \exp \left( - \frac{(2 h_i -1)^2  \rho_*^2  }{200} \frac{Tw}{Ln}(1 + o(1))\right).
$$
\ep
\end{proof}

\subsection{Adaptive selection strategy}

\begin{algorithm}
	\caption{Adaptive Clustering Algorithm. }
	\label{alg:adaptive_sampling}
	\begin{algorithmic}
		\STATE \textbf{Input: } $K$, $T$
		\STATE \textbf{Initialization: } $\tau \gets \left\lfloor {T} /\left( 4 \log\left(\frac{T}{n}\right) \right)\right\rfloor$
		\STATE $\hat{p}_{k \ell } \gets 0 $ for all $\ell \in [L], k \in [K]$
		\STATE $\hat{h}_i \gets 0.5$ for all $i \in \set{I}$
		\WHILE {$t < T$}
		\STATE $ Y_{i \ell} \gets \sum_{t'=1}^t \mathbbm{1}[i \in \set{W}_{t'}, \ell_{t'}= \ell]$
		\STATE $\hat{q}_{i \ell} \gets \frac{\sum_{t'=1}^t \mathbbm{1}[i \in \set{W}_{t'}, \ell_{t'}= \ell, X_{i \ell t'} = 1]}{Y_{i \ell}} $ if $Y_{i \ell}>0$, $\frac{1}{2}$ otherwise
		\IF {\textnormal{mod}$(t, \tau) = 0$}
		\STATE Use \text{Algorithm~\ref{alg:kmeans_for_r}} with input $K, t$ to  obtain the estimated clusters $\{\set{S}_k\}_{k = 1,\ldots,K}$
		\STATE {\bf Estimate statistical parameters:}
		\STATE	 $\hat{p}_{k \ell } \gets  \frac{1}{2} \left( \frac{\sum_{i \in \set{S}_k} (2 \hat{q}_{i \ell} - 1)}{|\set{S}_k|} + 1\right)$ 
		for all $\ell \in [L], k \in [K]$		
		\STATE { $\hat{h}_i \gets 
			\max\left\{0.5, \argmin_{h' \in [1/2, 1]} \left\{ \sum_{\ell \in [L]} Y_{i \ell} \KL (h' \hat{p}_{\hat{\sigma}(i) \ell} +\bar{h'} \bar{ \hat{p}}_{\hat{\sigma}(i) \ell}, \hat{q}_{i \ell}) \right\} - \sqrt{\frac{\log(t)}{ 10 \sum_{\ell } Y_{i \ell}}} \right\},$ 
		}
		\STATE for all $i \in \set{I}$ 
		\ENDIF
		\STATE {\bf Adaptive item question selection:} 
		\STATE $i^\star \gets \argmin_{i \in \set{I}} \hat{d}_i$	where $\hat{d}_{i} =  {d}_{i}(h'_i, k'_i, \vec{Y})$,
		\STATE where	$k'_{i} = \argmin_{k' \neq \hat{\sigma}(i)} \min_{h' \in [1/2, 1]} {d}_{i}(h', k',\vec{Y})$,
		\STATE $h'_{i} =  \argmin_{h' \in [1/2, 1]} {d}_{i}(h', k'_i, \vec{Y})$, 
		\STATE and ${d}_{i}(h', k', \vec{Y})
		= \sum_{\ell =1}^L Y_{i\ell} \KL ( 
		{h}' \hat{p}_{k' \ell}+\bar{{h}}' \bar{\hat{p}}_{k'\ell}, \hat{h}_i\hat{p}_{\hat{\sigma}(i) \ell}+\bar{\hat{h}}_i \bar{\hat{p}}_{\hat{\sigma}(i)\ell}) .$
		\STATE $\ell^\star \gets \argmax_{\ell} \min_{k' \neq \hat{\sigma}(i^\star)} 
		\KL(h'_{i^\star} \hat{p}_{k' \ell}  + h'_{i^\star} \bar{\hat{p}}_{k' \ell}, \hat{h}_{i^\star} \hat{p}_{\hat{\sigma}(i^\star) \ell} +\bar{\hat{h}}_{i^\star} \bar{\hat{p}}_{\hat{\sigma}(i^\star) \ell} ) $.
		\STATE Present items $\set{W}_t$ (including $i^\star$) with the smallest $\hat{d}_i$ with question $\ell^\star$.
		\ENDWHILE
		\STATE {\bf Output:}  $\{\set{S}_k\}_{k \in [K]}$
	\end{algorithmic}
\end{algorithm}

Our adaptive (item, question) selection and clustering algorithm is described in Algorithm \ref{alg:adaptive_sampling}. The design of the adaptive (item, question) selection strategy is inspired by the derivation of the information-theoretical error lower bounds. The algorithm maintains estimates of the model parameters $\vec{p}$ and $\vec{h}$ and of the clusters $\{\set{I}_k\}_{k\in [K]}$. These estimates, denoted by $\hat{\vec{p}}$, $\hat{\vec{h}}$, and $\{\set{S}_k\}_{k\in[K]}$, respectively, are updated every $\tau=T/(4\log(T/n))$ users. More precisely, we use Algorithm 1 to compute $\{\set{S}_k\}_{k\in [K]}$, and from there, we update the estimates as:
\begin{linenomath}
\begin{align*}
\hat{p}_{k \ell } & = \frac{1}{2} \left( \frac{\sum_{i \in \set{S}_k} (2 \hat{q}_{i \ell} - 1)}{|\set{S}_k|} + 1\right),
\\
\hat{h}_i & = \argmin_{h' \in [1/2, 1]} \left\{ \sum_{\ell \in [L]} Y_{i \ell} \KL (h' \hat{p}_{\hat{\sigma}(i) \ell} +\bar{h'} \bar{ \hat{p}}_{\hat{\sigma}(i) \ell}, \hat{q}_{i \ell}) \right\},
\end{align*}
\end{linenomath}
where $Y_{i\ell}$ is the number of times where question $\ell$ has been asked for item $i$ so far, and where $\hat{\sigma}(i)$ corresponds to the estimated cluster of $i$ (i.e., $i\in \set{S}_{\hat{\sigma}(i)}$). Let $\vec{Y}:= (Y_{i \ell})_{i\in \set{I}, \ell \in [L]}$. 

Now using the same arguments as those used to derive error lower bounds, we may estimate that after seeing the $t$-th user, a lower bound on the mis-classification error for item $i$ is $\exp\left(- \hat{d}_{i}(\vec{Y}) \right)$, where
\begin{linenomath}
\begin{align*}
\hat{d}_{i}(\vec{Y})  := \min_{k' \neq \hat{\sigma}(i)} \min_{h' \in [1/2, 1]}   \sum_{\ell =1}^L  Y_{i\ell} \KL ( 
{h}' \hat{p}_{k' \ell}+\bar{{h}}' \bar{\hat{p}}_{k'\ell},
\hat{h}_i\hat{p}_{\hat{\sigma}(i) \ell}+\bar{\hat{h}}_i \bar{\hat{p}}_{\hat{\sigma}(i)\ell}) \;.
\end{align*}
\end{linenomath}
\updated{The above lower bounds are heuristic in nature, as they are based solely on estimated parameters and clusters. These are derived from the divergence $ {\set{D}}_{\set{M}}^A (i, \vec{y})$ using \eqref{eq:divergence_adaptive},  with a particular emphasis on the adjustable parameters for item $i$.  
This approach takes a pessimistic view of the hardness parameters, with the exception of those for item $i$.
Revisiting the scenario of Example \ref{ex:1}, there is only one question ($L=1$) and the adaptability of the algorithm is principally determined by how the budget $T$ is allocated among the items.
Observe that, when $h_i$ is estimated to be small, the value of $\KL (
{h}' \hat{p}_{k’ \ell}+\bar{{h}}' \bar{\hat{p}}_{k’\ell},
\hat{h}_i\hat{p}_{\hat{\sigma}(i) \ell}+\bar{\hat{h}}_i \bar{\hat{p}}_{\hat{\sigma}(i)\ell})$ tends to be small. Conversely, when $h_i$ is estimated to be large, the value of $\KL (
{h}' \hat{p}_{k’ \ell}+\bar{{h}}' \bar{\hat{p}}_{k’\ell},
\hat{h}_i\hat{p}_{\hat{\sigma}(i) \ell}+\bar{\hat{h}}_i \bar{\hat{p}}_{\hat{\sigma}(i)\ell})$ tends to be large. 
Therefore, the more difficult the item $i$ is, the greater the need for a larger $Y_{i 1}$, and the higher the frequency of it being selected.}
Analyzing the accuracy of these lower bounds is particularly challenging (it is hard to analyze the estimated item hardness $\hat{h}_i$). Using these estimated lower bounds, we select the items and the question to be asked next. We put in the list $\set{W}_t$ the $w$ items with the smallest $\hat{d}_{i}(\vec{Y})$. The question $\ell$ is chosen to maximize the term:
$
\min_{k' \neq \hat{\sigma}(i^\star)} 
\KL(h'_{i^\star} \hat{p}_{k'\ell}  + h'_{i^\star} \bar{\hat{p}}_{k'\ell}, \hat{h}_{i^\star} \hat{p}_{\hat{\sigma}(i^\star) \ell} +\bar{\hat{h}}_{i^\star} \bar{\hat{p}}_{\hat{\sigma}(i^\star) \ell} ),
$
where $i^\star = \argmin_{i \in \set{I}} \hat{d}_i(\vec{Y})$ (see Algorithm 2 for the details).
Note that the question is selected by considering the item $i^\star$ that seems to be the most difficult to classify. 
 
 Note that in order to reduce the computational complexity of the algorithm, we may replace the KL function in the definition of  ${d}_i$ by a simple quadratic function (as suggested in the proof of Proposition~\ref{prop:bound-D-M}). This simplifies the minimization problem over $h'$ to find $h_i'$. We actually have an explicit expression for $h_i'$ with this modification.
 
The computational complexity of the adaptive algorithm (Algorithm~2 in Appendix) is: $\set{O}(n^2T/\tau) = \set{O}(n^2 \log(T/n))$. As in the uniform case, by choosing a small ($\log n$) subset of items (and not all the items in $\set{I}$) to compute centroids ($T_i$), one can reduce the computational complexity to: $\set{O}(n \log (n) \log (T/n)).$ We provide experimental evidence on the superiority of our adaptive algorithm in the following sections.

%% file: Numerical_experiments.tex
\section{Numerical experiments: Synthetic data}
In this section, we evaluate the performance of our algorithms on synthetic data. We consider different models.  the problem investigated here is different from those one may find in the  crowdsourcing or Stochastic Block Model literature. Hence, we cannot compare our algorithms to existing algorithms developed in this literature. Instead we focus on comparing the performance of our nonadaptive and adaptive algorithms.

\begin{figure}[ht]
	\centering
	\begin{minipage}[b]{0.7\linewidth}
		\centering
		\includegraphics[width=0.8\columnwidth]{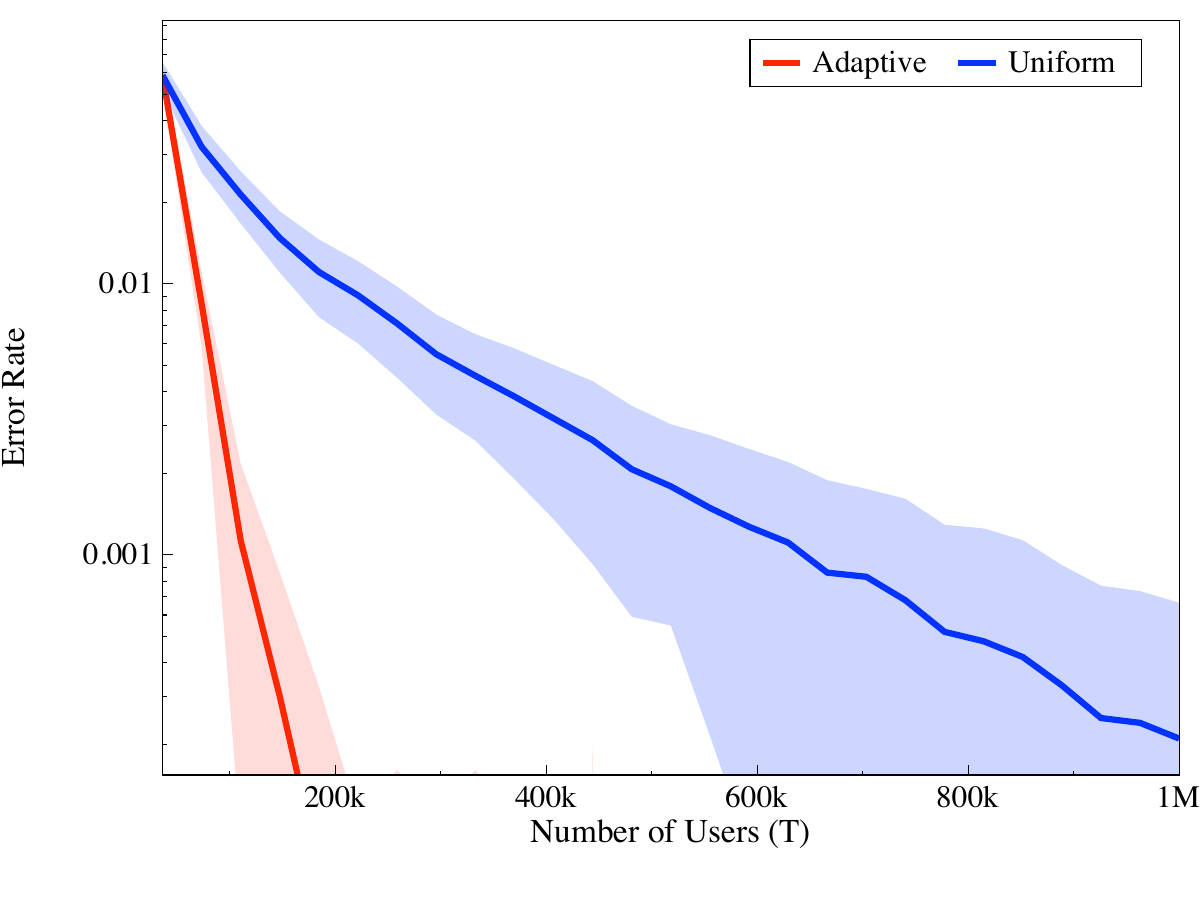} 
		\end{minipage}
	\\
		\begin{minipage}[b]{0.7\linewidth}
			\centering
			\includegraphics[width=0.8\columnwidth]{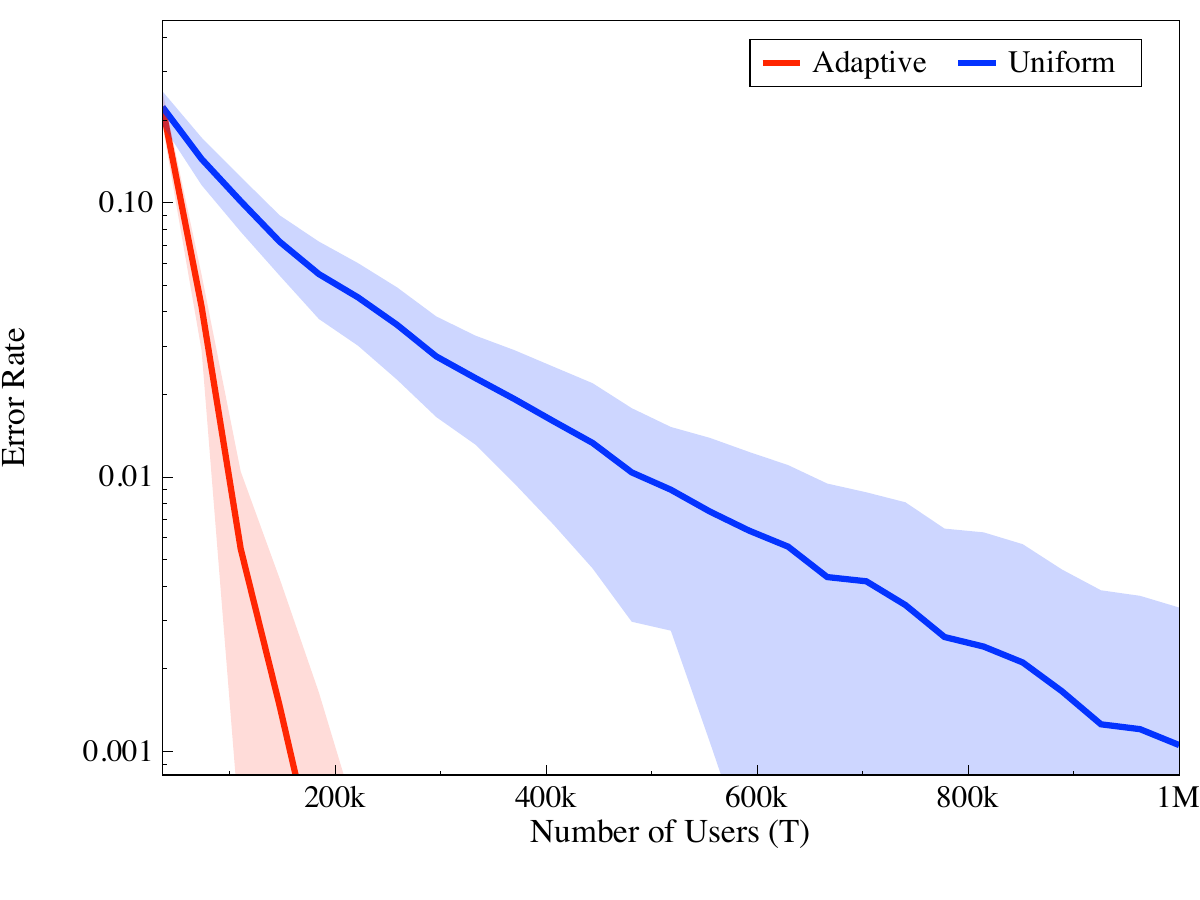} 
	\end{minipage}
\caption{Model 1. (top) Global error rate vs. number of users. (bottom) Error rate for the 20\% most difficult items vs. number of users. One standard deviations are shown using shaded areas.} \label{fig:error_rate_0_100}
\end{figure}

\begin{figure}[ht]
	\centering
	\includegraphics[width=0.6\columnwidth]{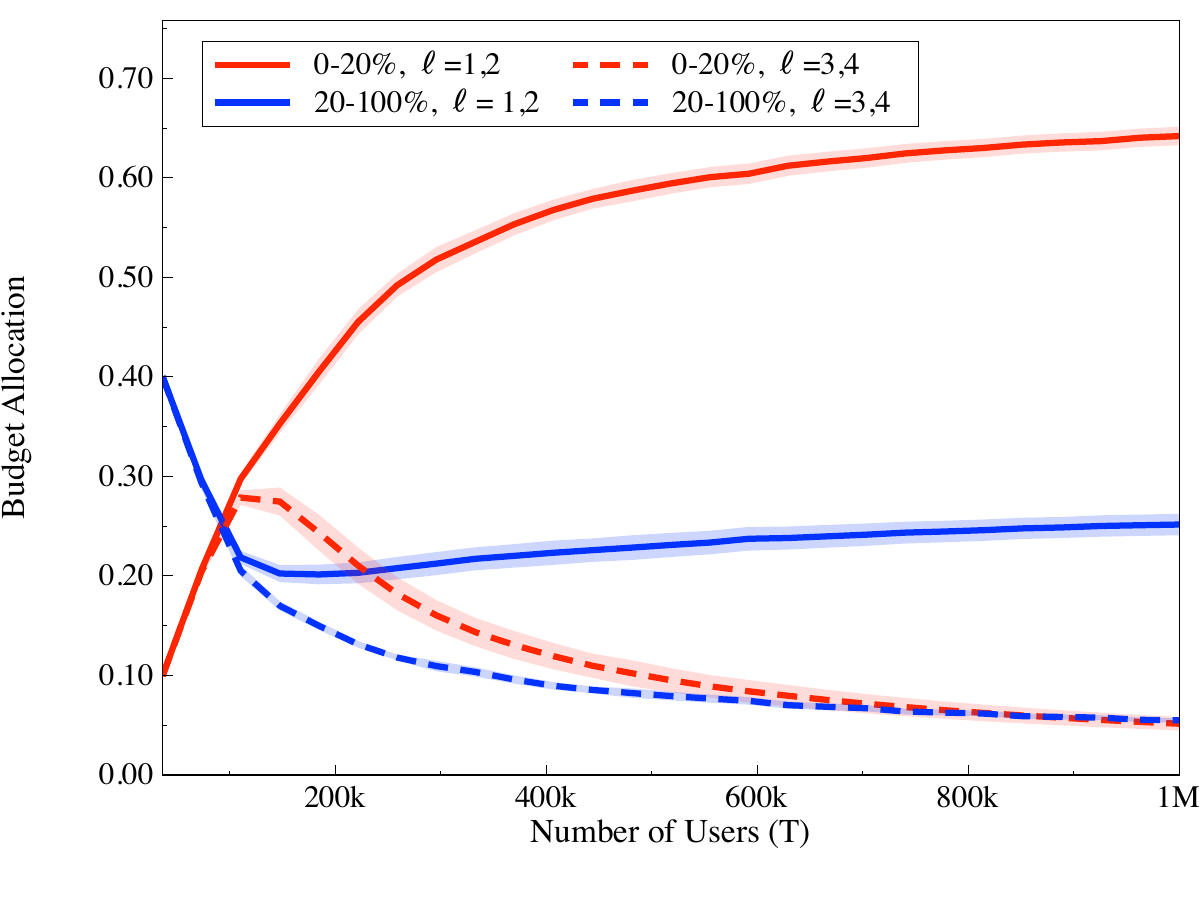}
	\caption{Model 1. The budget allocation under the adaptive algorithm vs. number of users. Items and questions are grouped into 4 categories, e.g. $(0-20\%, \ell=1,2)$ is the category regrouping the 20\% most difficult items and questions $\ell=1,2$. One standard deviations are shown using shaded areas.
	}
	\label{fig:budget_diff}
\end{figure}

\medskip
\noindent {\bf Model 1 - heterogeneous items with dummy questions. } Consider $n=1000$ items and two clusters ($K=2$) of equal sizes. The hardness of the items are i.i.d., picked uniformly at random in the interval $[0.55, 1]$. We ask each user to answer one of four questions. The answers' statistics are as follows: for cluster $k =1$, $\vec{p}_{1} = (0.01, 0.99, 0.5, 0.5)$
and for cluster $k =2$, $\vec{p}_{2} = (0.99, 0.01, 0.5, 0.5)$. Note that only
half of the questions ($\ell = 1, 2$) are useful; the other questions ($\ell = 3,4$) generate completely random answers for both clusters. 
Figure~\ref{fig:error_rate_0_100} (top) plots the error rate averaged over all items and over $100$ instances of our algorithms.
Under both algorithms, the error rate decays exponentially with the budget $T$, as expected from the analysis. Selecting items and questions in an adaptive manner brings significant performance improvements. For example, after collecting the answers from $t=200k$, the adaptive algorithm recovers the clusters exactly for most of the instances, whereas the algorithm using uniform selection does not achieve exact recovery even with $t=1000k$ users. In particular, the adaptive algorithm is able to reduce the error rates on the 20\% most difficult items, i.e., items that have the $20\%$ smallest $h_i$. In Figure~\ref{fig:error_rate_0_100} (bottom), we present the error rate of these items. The error rates for these most difficult items are significantly reduced by being adaptive. In Figure~\ref{fig:budget_diff}, we present the evolution over time of the budget allocation observed under our adaptive algorithm. We group items and questions into four categories. 
For example, one category corresponds to the question $\ell=1,2$ and to the 20\% most difficult items. As expected, the adaptive algorithm learns to select relevant questions ($\ell=1,2$) with hard items more and more often as time evolves. 

\begin{figure}[ht]
	\centering
	\includegraphics[width=0.6\columnwidth]{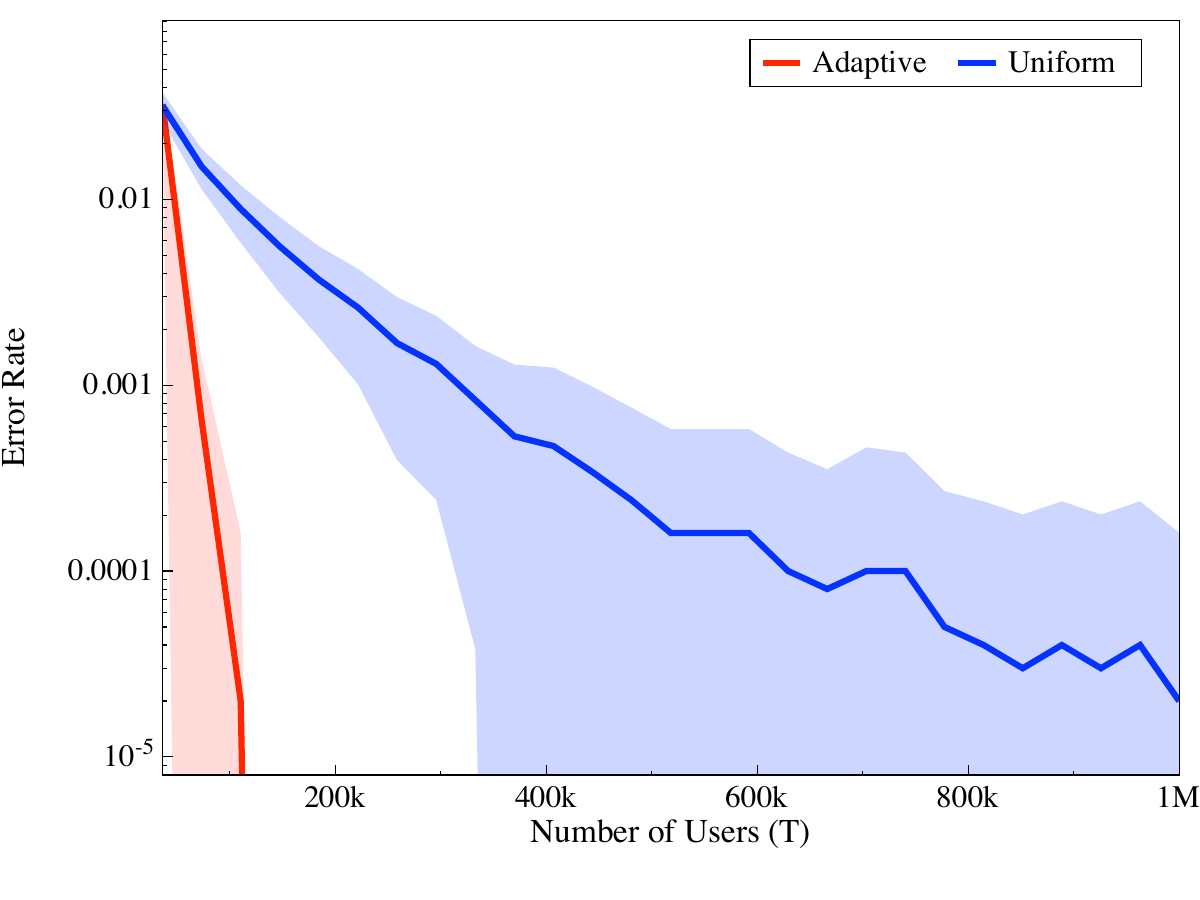}
	\caption{Model 2. Global error rate vs. number of users. One standard deviations are shown using shaded areas. }
	\label{fig:nodummy}
\end{figure}

\medskip
\noindent {\bf Model 2 - heterogeneous items without dummy questions. } This model is similar to Model 1, except that we remove the dummy questions $\ell=3, 4$, i.e., we set $\vec{p}_{1} = (0.01, 0.99)$ and $\vec{p}_{2} = (0.99, 0.01)$. The performance of our algorithms are shown in Figure~\ref{fig:nodummy}. Overall, compared to Model 1, the error rates are better. For example, exact cluster recovery is achieved using only $100k$ users for almost all instances.

\begin{figure}[ht]
	\centering
	\includegraphics[width=0.6\columnwidth]{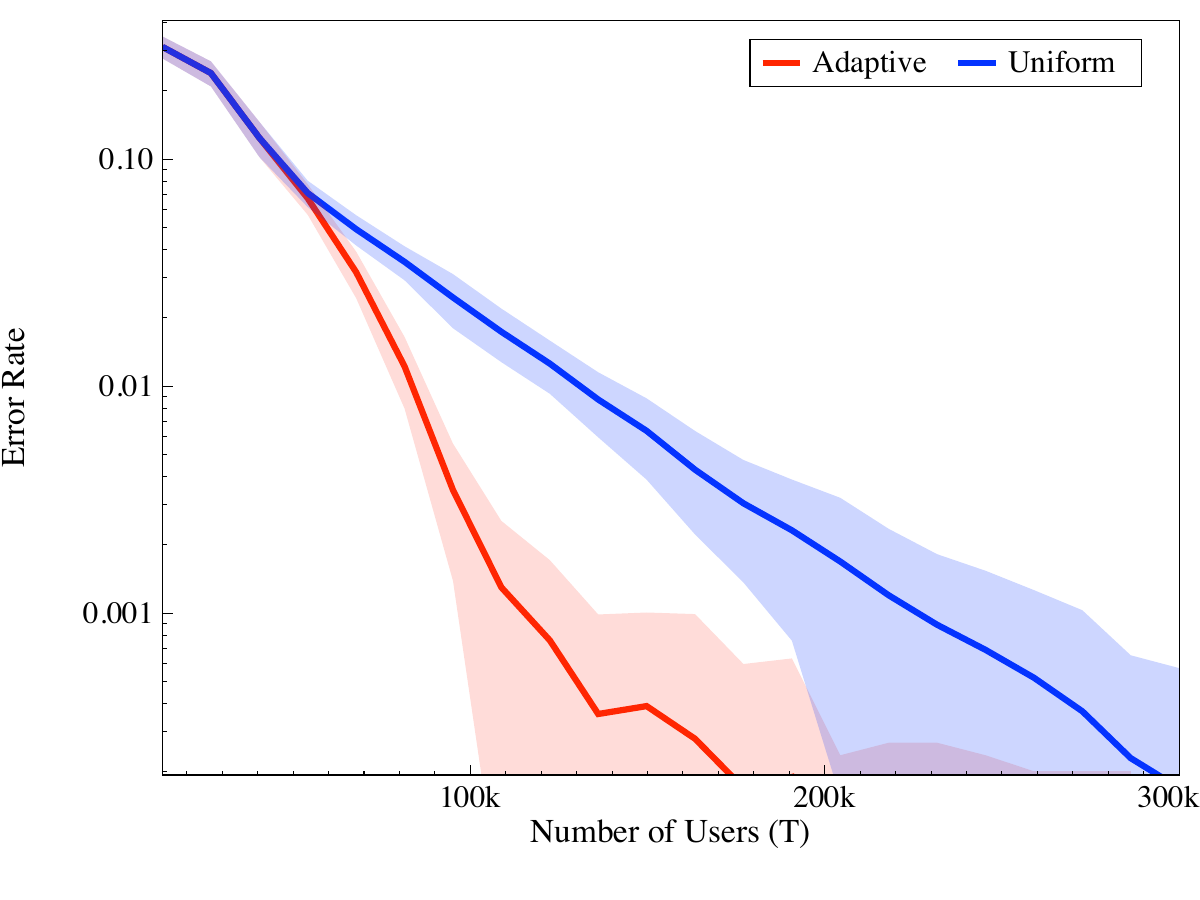}
	\caption{Model 3. Global error rate vs. number of users. One standard deviations are shown using shaded areas.}
	\label{fig:nodegcor}
\end{figure}

\medskip
\noindent {\bf Model 3 - homogeneous items with dummy questions.} Here we study the homogeneous scenario where all items have the same hardness: $h_i = 1, \forall i \in \set{I}$. We still have $1000$ items grouped into two clusters of equal sizes. We set $\vec{p}_{1} = (0.3, 0.2,0.2,0.2)$, $\vec{p}_{2} = (0.7, 0.2, 0.2,0.2)$ (question $\ell = 2, 3, 4$ are useless). The performance of the algorithms is shown in Figure~\ref{fig:nodegcor}. The adaptive algorithm exhibits better error rates than the algorithm with uniform selection, although the improvement is not as spectacular as in heterogeneous models where adaptive algorithms can gather more information about difficult items. In homogeneous models, the adaptive algorithm remains better because it selects questions wisely.

\begin{figure}[ht]
	\centering
	\includegraphics[width=0.6\columnwidth]{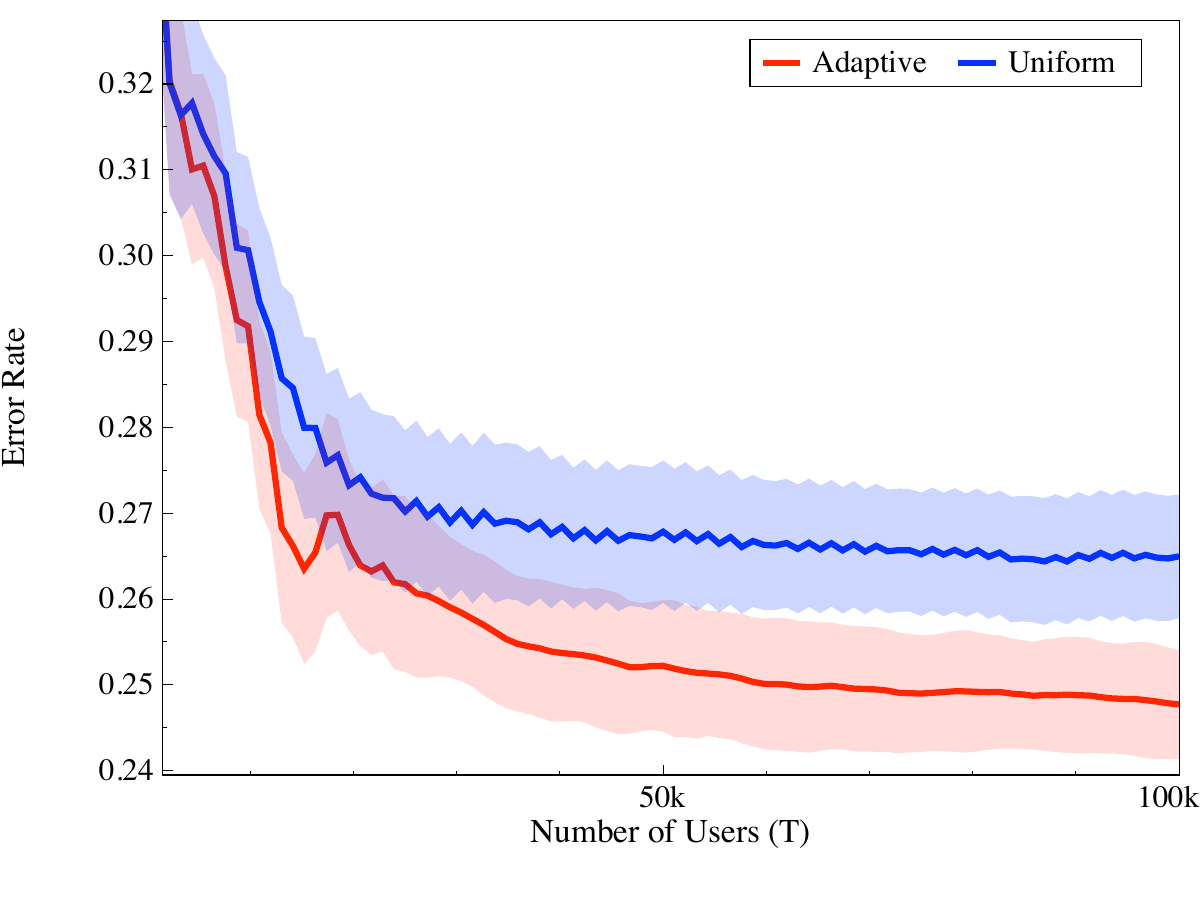}
	\caption{Real-World data:  global error rate vs. number of users. One standard deviations are shown using shaded areas.	
	}
	\label{fig:Realworld_errorrate}
\end{figure}

%% file: nonsynthetic.tex
\section{Numerical experiments: Real-world data}
Finally, we use real-world data to assess the performance of our algorithms. Finding data that would fit our setting exactly (e.g. several possible questions) is not easy. We restrict our attention here to scenarios with a single question, but with items with different hardnesses. We use the waterbird dataset by \cite{welinder2010multidimensional}. This dataset contains 50 images of Mallards (a kind of duck) and 50 images of Canadian Goose (not a duck). The dataset reports the feedback of 40 users per image, collected using Amazon Mturk: each user is asked whether the image is that of a duck. 
\updated{This scenario mirrors the one outlined in Example~\ref{ex:1} in Introduction.}
Each image is unique in the sense that the orientation of the animal varies, the brightness and contrasts are different, etc. We hence have a good heterogeneity in terms of item hardness. Actually, the classification task is rather difficult, and the users' answers seem very noisy -- overall answers are correct 76\% of the time.

From this small dataset, we generated a larger dataset containing 1000 images (by just replicating images). To emulate the sequential nature of our clustering problem, in each round, we pick a user uniformly at random (with replacement), and observe her answers to the selected images. 

The error rates of both algorithms are shown in Figure~\ref{fig:Realworld_errorrate}. The global error rate is averaged over 100 instances. Both algorithms have rather low performance, which can be explained by the inherent hardness of the learning task. The adaptive algorithm becomes significantly better after $t=20k$ users. this can be explained as follows. The adaptive algorithm needs to estimate the hardness of items before being efficient. Until the algorithm gathers enough answers on item $i$, its estimate of $\hat{h}_i $ remains close to $0.5$. As a consequence, the algorithm continues to pick items uniformly at random. As soon as the algorithm gets better estimates of the items' hardnesses, it starts selecting items with strong preferences.

%% file: Conclusion.tex
\clearpage
\section{Conclusion}

In this paper, we analyzed the problem of clustering complex items using very simple binary feedback provided by users. A key aspect of our problem is that it takes into account the fact that some items are inherently more difficult to cluster than some others. Accounting for this heterogeneity is critical to get realistic models, and is unfortunately not investigated often in the literature on clustering and community detection (e.g. that on Stochastic Block Model). The item heterogeneity also significantly complicates any theoretical development. 

For the case where data is collected uniformly (each item receives the same amount of user  feedback), we derived a lower bound of the clustering error rate for any individual item, and we developed a clustering algorithm approaching the optimal error rate. We also investigated adaptive algorithms, under which the user feedback is received sequentially, and can be adapted to past observations. Being adaptive allows to gather more feedback for more difficult items. We derived a lower bound of the error rate that holds for any adaptive algorithm. Based on our lower bounds, we devised an adaptive algorithm that smartly select items and the nature of the feedback to be collected. We evaluated our algorithms on both synthetic and real-world data. These numerical experiments support our theoretical results, and demonstrate that being adaptive leads to drastic performance gains.

%% file: app_table.tex
\section{Table of notations}

\begin{table}[htbp]
	\begin{center}%
		\small 
		\begin{tabular}{c c p{10cm} }
			\toprule
			\multicolumn{3}{l}{\bf Problem-specific notations}\\
			\hline
			$\set{I}$ & $ $ & Set of items\\
			$\set{I}_k$ & $ $ & Set of items in the item cluster $k$\\
			$n $ & $ $ & Number of items\\
			$ K$ & $ $ & Number of item clusters\\ 
			$\sigma(i)$& & Cluster index of item $i$\\
			$\vec{\alpha} \coloneqq (\alpha_1, \ldots, \alpha_K)$ & & Fractions of items in each cluster \\
			$ w$ & &  The number of items presented at the same time\\
			$ \set{W}_t$ & & Set of items presented to the $t$-th user\\
			$ L$ & & Number of possible questions\\
			$ \ell_t $  & & Question asked to the $t$-th user\\
			$T$ & & Total number of user arrivals within the time horizon\\
			$\vec{p} \coloneqq (p_{k \ell})_{k\in [K], \ell \in [L]}$& & Statistical parameterization of items in cluster $k$ for the question $\ell$\\
			$\vec{h} \coloneqq (h_i)_{i \in \set{I}}$ & & Hardness parameter for item $i$\\
			$ \set{M}$& & Statistical models parameterized by $(\vec{p}, \vec{h})$\\
			$ X_{i \ell t} $ && Binary feedback from $t$-th user for item $i$ and for question $\ell$\\
			$q_{i \ell} \coloneqq h_i p_{k \ell} - \bar{h}_i \bar{p}_{k \ell} $ & & Probability of positive answer to item $i$ for question $\ell$\\
			$h_\star$ & & Minimum hardness across items, see Assumption (A1)\\
			$\rho_\star$ & & Minimum separation between different clusters, see Assumption (A1)\\
			$ \eta$& & Homogeneity parameter among clusters, see Assumption (A2)\\
			$\Omega$ & & Set of all models satisfying Assumptions (A1) and (A2)\\
			$r_{k\ell}$& & Value of $ 2 p_{k \ell} -1$\\
			$\set{E}^\pi$ & & Set of misclassified items by the algorithm $\pi$\\
			$\varepsilon^\pi_i (n, T)$ & & Probability that the item $i$ is misclassifed after the $T$-th user arrived under the algorithm $\pi$\\
			$ \varepsilon^\pi(n, T)$ & & Expected proportion of misclassified items after the $T$-th user arrived under the algorithm $\pi$\\
			$Y_{i \ell}$& & Number of times the item $i$ is presented together with the question $\ell$\\
			$y_{i \ell}$& & Normalized expected number of times the question $\ell$ is asked for the item $i$ under some fixed algorithm\\
			$\set{D}^U_{\set{M}}(i)$& & Divergence for the misclassification of item $i$ with the model $\set{M}$ under uniform item selection strategy\\
			$\tilde{\set{D}}^U_{\set{M}}$& & Global divergence with the model $\set{M}$ under uniform item selection strategy\\
			$\set{D}^A_{\set{M}}(i, \vec{y})$& & Divergence for the misclassification of item $i$ with the model $\set{M}$ under some adaptive item selection strategy satisfying $\EXP[Y_{i \ell}] = \frac{Tw}{n} y_{i \ell}$\\
			$\tilde{\set{D}}^A_{\set{M}}$& & Global divergence with the model $\set{M}$ under the optimal adaptive item selection strategy\\
		\multicolumn{3}{c}{}\\
		\hline
        \end{tabular}
	\normalsize
	\end{center}
	\label{tab:TableOfNotations}
\end{table}

\begin{table}[htbp]
	\begin{center}%
		\small 
		\begin{tabular}{c c p{10cm} }
			\toprule
		\multicolumn{3}{l}{\bf Generic notations}\\
		\hline
		$\hat{a}$& & Estimated value of $a$\\
		$[a]$& & Set of positive integers upto $a$, i,e., $\{1, \ldots, a\}$\\
		$\mathbbm{1}\{A\}$ & & Indicator function: $0$ when $A$ is false, $1$ when $A$ is true \\
		$ \|\vec{x}\|$ & & $\ell_\infty $ norm of $\vec{x}$, i.e., $\|\vec{x}\| = \max_i x_i$\\
		$\|\vec{x}\|_2$ & & $\ell_2 $ norm of $\vec{x}$\\
		$\bar{a}$ & & Value of $1 - a$\\
		$\Pr(A)$ & & Probability that event $A$ occurs\\
		$ \EXP[a]$ & & Expected value of $a$\\
		$\kl (a, b)$ & & Kullback-Leibler divergence between Bernoulli distributions with means $a $ and $b$
			\\ \bottomrule
		\end{tabular}
	\normalsize
	\end{center}
	\label{tab:TableOfNotations2}
\end{table}

\clearpage

%% file: appendix.tex
\input{lower-proof}

\section{Proof of Lemma \ref{lm:concentration_of_r}}\label{sec:prooflems}

We use Hoeffding's inequality to establish the lemma.

\begin{theorem}[Hoeffding's inequality for bounded independent random variables (Theorem 1 of \cite{hoeffding1963probability})] Let $X_1,\dots,X_n$ be independent random variables with values in $[0,1]$. Denote $\mu = \EXP\left[ \frac{1}{n}\sum_{i=1}^n X_i\right].$ Then, for any $t \geq 0$,
\begin{align*}
	\mathbb{P} \left\{\frac{1}{n} \sum_{i=1}^n X_i -\mu \ge t \right\} & \leq \exp\left( - 2 n t^2 \right) \;.
\end{align*}
\label{thm:hoeffdings}
\end{theorem}
 
\begin{lemma} 	\label{lm:concentration_of_qi_by_chernoff}
Recall that by definition, $\tau := \lfloor \frac{T w}{L n} \rfloor$. For any $\varepsilon > 0$, $\|\hat{\vec{q}}_{i} - \vec{q}_{i} \| \leq \varepsilon $ with probability at least $1 - 2 L \exp \left(- {2 \tau}\varepsilon^2\right)\;.$
\end{lemma}
\begin{proof}[\updated{Proof of Lemma~\ref{lm:concentration_of_qi_by_chernoff}}]
    Note that the number of times question $\ell$ is asked for item $i$ is $\tau$. Using Hoeffding's inequality (Theorem \ref{thm:hoeffdings}), 
it is straightforward to check: for any $\varepsilon > 0$ and $\ell \in [L]$,
\begin{align*}
\mathbb{P}\left\{\left|\frac{1}{\tau}\sum_{t=1}^{\tau} (X_{i \ell t} -  q_{i \ell })\right|
=| \hat{q}_{i \ell } - q_{i \ell }|
 \geq   {\varepsilon}\right\} \leq 2 \exp \left(-   2 \tau  \varepsilon^2  \right).\end{align*}
We conclude the proof using the union bound as follows:
\begin{align*}
\mathbb{P}\left\{ \| \hat{\vec{q}}_{i} -   {\vec{q}}_{i} \| \geq \varepsilon \right\}
&
\le
\sum_{\ell \in [L]}\mathbb{P}\left\{ | \hat{{q}}_{i\ell} -   {{q}}_{i\ell} | \geq {\varepsilon}
\right\}
\leq 2 L \exp \left( - 2 \tau \varepsilon^2  \right).
\end{align*}
\ep
\end{proof}

\begin{proof}[\updated{Proof of Lemma~\ref{lm:concentration_of_r}}]
     By Lemma \ref{lm:concentration_of_qi_by_chernoff}, we have $\| \hat{\vec q}_i - {\vec q}_i \| \leq \varepsilon$ with probability at least $1 - 2 L \exp \left(- {2 \varepsilon^2} \tau \right)$. Suppose $\| \hat{\vec q}_i - {\vec q}_i \| \leq \varepsilon$ and $0 < \varepsilon \le \frac{\|2 \vec{q}_i - 1\|}{16}$. Using the triangle inequality and the reverse triangle inequality, we have 
\begin{align*}
\| 2 {\vec q}_i - 1\| - 2 \| {\vec q}_i - \hat{\vec q}_i \|  \leq \|2 \hat{\vec q}_i - 1  \| \leq \| 2 {\vec q}_i - 1 \| + 2\| {\vec q}_i - \hat{\vec q}_i \|,
\end{align*}
Therefore, 
\begin{align*}
\| 2 {\vec q}_i - 1\| - 2 \varepsilon \leq \|2 \hat{\vec q}_i - 1  \| \leq \| 2 {\vec q}_i - 1 \| + 2\varepsilon
\end{align*}
which implies that:
\begin{align*}
\frac{1}{\| 2 {\vec q}_i - 1\|} \frac{1}{1+ \frac{2 \varepsilon}{\| 2 {\vec q}_i - 1\| }}  \leq \frac{1}{\|2 \hat{\vec q}_i - 1  \|} \leq \frac{1}{\| 2 {\vec q}_i - 1\|} \frac{1}{1 -  \frac{2 \varepsilon}{\| 2 {\vec q}_i - 1\| }}.
\end{align*}
From $0 < \varepsilon \le \frac{\|2 \vec{q}_i - 1\|}{16},$ we have $0< \frac{2 \varepsilon}{\|2 \vec{q}_i - 1\|} \leq \frac{1}{8}$. Now observe that we have:
\begin{align*}
1- x \leq \frac{1}{1+x} \quad \text{and} \quad \frac{1}{1-x} \leq 1 + \frac{8}{7}x \;,
\end{align*}
for all $x$ such that $0<x< \frac{1}{8}$. Then we obtain:
\begin{align*}
\frac{1}{\| 2 {\vec q}_i - 1\|}\left(1 - \frac{2 \varepsilon}{\| 2 {\vec q}_i - 1\|} \right) \leq \frac{1}{\| 2 \hat{\vec q}_i - 1\|} \leq \frac{1}{\| 2 {\vec q}_i - 1\|}\left(1 + \frac{8}{7}\frac{2 \varepsilon}{\| 2 {\vec q}_i - 1\|}  \right).
\end{align*}
Then, there exists $x \in \left[- \frac{2 \varepsilon}{\| 2 {\vec q}_i - 1\|} ,  \frac{16 \varepsilon}{7 \| 2 {\vec q}_i - 1\|} \right]$ such that $ \frac{1}{\| 2 \hat{\vec q}_i - 1\|} = \frac{1}{\| 2 {\vec q}_i - 1\|} (1 + x)$. Using this $x$, we get:
\begin{align*}
\| \hat{\vec r}^i - \tilde{\vec r}^i \| & = \left\| \frac{2 \hat{\vec q}_i -1 }{\|2 \hat{\vec q}_i -1  \|} - \frac{2 {\vec q}_i -1 }{\|2 {\vec q}_i -1  \|} \right\|
\\
& = \frac{1}{\|2 {\vec q}_i -1  \|} \left\| \frac{\|2 {\vec q}_i -1  \|}{\|2 \hat{\vec q}_i -1  \|}(2 \hat{\vec q}_i -1 ) -  (2 {\vec q}_i - 1)\right\|
\\
& = \frac{1}{\|2 {\vec q}_i -1  \|} \left\| (1 + x) (2 \hat{\vec q}_i -1 ) -  2 {\vec q}_i + 1 \right\|
\\
&
\leq  \frac{1}{\|2 {\vec q}_i -1  \|}  \left( \left\| 2 (\hat{\vec q}_i - \vec{q}_i)\right\| + \left\| x (2 \hat{\vec q}_i - 1) \right\|\right)
\\
&
\leq  \frac{1}{\|2 {\vec q}_i -1  \|}  \left( \left\| 2 (\hat{\vec q}_i - \vec{q}_i)\right\| + \left\| x (2 \hat{\vec q}_i - 2 \vec{q}_i) \right\| + \|x(2 \vec{q}_i -1) \| \right)
\\
&
\leq  \frac{1}{\|2 {\vec q}_i -1  \|} (2 \varepsilon + \frac{16}{7}\varepsilon \cdot 2 \varepsilon + \frac{16}{7} \varepsilon)
\\
&
=  \frac{1}{\|2 {\vec q}_i -1  \|} (\frac{30}{7}\varepsilon  + \frac{32}{7} \varepsilon^2)
\\
&
\stackrel{(a)}{\leq} \frac{5 \varepsilon }{\|2 {\vec q}_i -1 \|},
\end{align*}
with probability at least $1 - 2 L \exp \left(- {2 \varepsilon^2} \tau \right)$ for all $\varepsilon$ such that $0< \varepsilon \leq \frac{\|2 \vec{q}_i - 1\|}{16}$, where in $(a)$, we use $5x \geq \frac{30}{7}x + \frac{32}{7} x^2 $ for all  $0<x<\frac{1}{16}$ and $0<\varepsilon \leq \frac{\|2 \vec{q}_i - 1 \|}{16} \leq \frac{1}{16}$ . 
This concludes the proof.
\ep
\end{proof}

%% file: lower-proof.tex
\section{Proof of Proposition~\ref{prop:bound-D-M}}
\label{sec:pf-bound-D-M}

For given $i \in \set{I}$,
let $k = \sigma(i)$ and $k' \in [K]$ be such that:
\begin{align*}
\set{D}_{\set{M}}^U(i) = 
\min_{h' \in [(h_*+1) /2, 1]}  \frac{1}{L} \sum_{\ell =1}^L \KL (h'p_{k'\ell} + \bar{h}'\bar{p}_{k'\ell},
q_{i\ell}) \;.
\end{align*}

{\bf Upper bound.} %
Recalling the definition of $q_{i\ell} := h_i p_{k\ell} + \bar{h}_i \bar{p}_{k\ell}$, it follows that for any $h' \in [(h_*+1) /2, 1],$
\begin{align*}
\set{D}_{\set{M}}^U(i) & \leq \frac{1}{L} \sum_{\ell =1}^L \KL (h' p_{k'\ell} + \bar{h}' \bar{p}_{k'\ell},
 h_i p_{k\ell} + \bar{h}_i \bar{p}_{k\ell} )   \\
 &\le  
\frac{1}{L} \sum_{\ell =1}^L 
((h' p_{k'\ell} + \bar{h}' \bar{p}_{k'\ell}) -  (h_i p_{k\ell} + \bar{h}_i \bar{p}_{k\ell}))^2
\left( \frac{1}{q_{i\ell}}
+\frac{1}{\bar{q}_{i\ell}}
\right)
 \\
  &\le  
\frac{2}{ L \eta } \sum_{\ell =1}^L 
((h' p_{k'\ell} + \bar{h}' \bar{p}_{k'\ell}) -  (h_i p_{k\ell} + \bar{h}_i \bar{p}_{k\ell}))^2
 \\
 &=
\frac{2}{ L\eta } \sum_{\ell =1}^L 
\left( \frac{(2h'-1)(2 p_{k' \ell} - 1) + 1}{2} - \frac{(2h_i-1)(2 p_{k \ell} - 1) +1}{2} \right)^2 \\
 &=
\frac{1}{ 2L\eta } 
(2h_i-1)^2 
\left\| \frac{2h' - 1}{2h_i - 1} \vec{r}_{k'} - \vec{r}_{k} \right\|^2_2 \;.
\end{align*}
where the second inequality is from the comparison between the KL divergence
from $\chi^2$-divergence
and the third inequality is from (A2), i.e., $q_{i\ell} \in [\eta, 1-\eta]$. 
Now observing that \updated{$h' \in [(h_*+1) /2, 1]$ implies $ \frac{h_*}{2h_i - 1} \leq \frac{2h' - 1}{2h_i - 1} \leq \frac{1}{2h_i - 1}.$} 
Taking the minimum over $h' \in [(h_*+1) /2, 1]$,
we obtain the upper bound.

{\bf Lower bound.}
Using Pinsker's inequality, we obtain:
\begin{align*}
\set{D}_{\set{M}}^U(i) 
& \ge 
\frac{2}{L} \sum_{\ell =1}^L 
\left| (h'p_{k'\ell} + \bar{h}'\bar{p}_{k'\ell})
- (h_i p_{k\ell} + \bar{h}_i \bar{p}_{k\ell}) \right|^2  \\
& =
\frac{1}{2L} \sum_{\ell =1}^L 
\left| (2h'-1)(2p_{k'\ell}-1)
- (2h_i-1)(2p_{k\ell}-1) \right|^2  \\
& = 
\frac{1}{2L} 
\left\|  (2h'-1)\vec{r}_{k'}
- (2h_i-1)  \vec{r}_{k} \right\|^2_2 \\
& = 
\frac{(2h_i-1)^2}{2L} 
\Big\|  \left(\frac{2h'-1}{2h_i-1} \right) \vec{r}_{k'} - \vec{r}_{k} \Big\|^2_2 \\
& \ge
\frac{(2h_i-1)^2}{2L} 
\min_{\updated{\frac{h_*}{2h_i - 1} \leq \frac{2h' - 1}{2h_i - 1} \leq \frac{1}{2h_i - 1}} }
\left\|  c \vec{r}_{k'} - \vec{r}_{k} \right\|^2_2 \;,
\end{align*}
where for the last inequality, \updated{we again use the fact that $h' \in [(h_*+1) /2, 1]$ implies
$ \frac{h_*}{2h_i - 1} \leq \frac{2h' - 1}{2h_i - 1} \leq \frac{1}{2h_i - 1}$.}
This completes the proof of Proposition~\ref{prop:bound-D-M}. Note that we can further write:
\begin{align*}
\set{D}_{\set{M}}^U(i) 
&\ge
\frac{1}{2L} 
(2h_i-1)^2
\min_{ \updated{\frac{h_*}{2h_i - 1} \leq \frac{2h' - 1}{2h_i - 1} \leq \frac{1}{2h_i - 1}}} \left\|  c\vec{r}_{k'} - \vec{r}_k  \right\|^2_2  \\
&\ge 
 \frac{1}{2L} 
 (2h_i-1)^2
\rho_*^2  \;,
\end{align*}
using the relationship between the $\ell_\infty$-norm and the Euclidean norm and (A1).
\ep

\section{Proof of Lemma~\ref{lem:random-log-calc}}
\label{sec:pf-random-log-calc}

$\mathbb{E}_{\set{N}} [L] $ can be obtained as follows:
\begin{align}  
\EXP_{{\set{N}}}[\set{L}] & 
= \sum_{\ell = 1}^L \sum_{t=1}^T \mathbbm{1}[i \in \set{W}_t \text{ and } \ell_t = \ell] \EXP_{{\set{N}}}\left[\mathbbm{1}[X_{i\ell t} = +1] \log \frac{q_{\ell}'}{q_{i \ell}} + \mathbbm{1}[X_{i\ell t} = -1] \log \frac{\bar{q}_\ell'}{ \bar{q}_{i \ell} }\right] \nonumber
\\
& = \sum_{\ell =1}^L \frac{Tw}{Ln} 
 \left(
q'_{\ell} \log \frac{q'_{\ell}}{q_{i\ell}}
+\bar{q}'_{\ell} \log \frac{\bar{q}'_{\ell}}{\bar{q}_{i\ell}}
\right) \nonumber
\\
& = \frac{Tw}{Ln} \sum_{\ell =1}^L \KL(q'_{\ell}, q_{i\ell}) ) \nonumber
\\
& = \frac{Tw}{n} \set{D}_{\set{M}}^U (i)\;.\label{eq:random-log-mean}
\end{align}

To bound the variance of $\set{L}$, we first decompose $\set{L}^2$ as follows:
\begin{align*} 
\set{L}^2 = 
\sum_{t=1}^T \set{L}_{t}^2
+ 
\sum_{t\neq t'} \set{L}_{t}\set{L}_{t'} \;,
\end{align*}
where $\set{L}_{t} :=
\mathbbm{1}[i \in {\set{W}}_t] 
\sum_{\ell =1}^L \mathbbm{1}[\ell_t = \ell]
 \left(
\mathbbm{1}[X_{i \ell t} = +1] \log \frac{q'_{\ell}}{q_{i\ell}}
+\mathbbm{1}[X_{i \ell t} = -1]  \log \frac{\bar{q}'_{\ell}}{\bar{q}_{i\ell}}
\right)
 $.
We compute $ \set{L}_{t}^2$ as follows:
 \begin{align}
 \set{L}_{t}^2 &=
\mathbbm{1}[i \in {\set{W}}_t] 
\sum_{\ell =1}^L \mathbbm{1}[\ell_t = \ell]
 \left(
\mathbbm{1}[X_{i \ell t} = +1] \log \frac{{q}'_{\ell}}{q_{i\ell}}
+\mathbbm{1}[X_{i \ell t} = -1]  \log \frac{\bar{q}'_{\ell}}{\bar{q}_{i\ell}}
\right)^2  \nonumber \\
& = 
\mathbbm{1}[i \in {\set{W}}_t] 
\sum_{\ell =1}^L \mathbbm{1}[\ell_t = \ell]
 \left(
\mathbbm{1}[X_{i \ell t} = +1] \left(\log \frac{q'_{\ell}}{q_{i\ell}} \right)^2
+\mathbbm{1}[X_{i \ell t} = -1]  \left( \log \frac{\bar{q}'_{\ell}}{\bar{q}_{i\ell}}\right)^2
\right) \nonumber  \\
& \le 
\log (1/\eta)
\mathbbm{1}[i \in {\set{W}}_t] 
\sum_{\ell =1}^L \mathbbm{1}[\ell_t = \ell]
 \left(
\mathbbm{1}[X_{i \ell t} = +1] \left|\log \frac{q'_{\ell}}{q_{i\ell}} \right|
+\mathbbm{1}[X_{i \ell t} = -1]  \left| \log \frac{\bar{q}'_{\ell}}{\bar{q}_{i\ell}}\right|
\right), \label{eq:calc-square-L}
\end{align}
where the last inequality follows from the fact that $q_{i \ell} \in [\eta, 1-\eta]$ under (A2), i.e., $\log \frac{q'_\ell}{q_{i\ell}} \le \log \frac{1}{\eta}$ and $\log \frac{\bar{q}'_\ell}{\bar{q}_{i\ell}} \le \log \frac{1}{\eta}$.
We deduce that:
\begin{align}
&\EXP_{\set{N}} \left[ \sum_{t=1}^T \set{L}_{t}^2 \right] = \sum_{t=1}^T \EXP_{\set{N}}\left[ \set{L}_{t}^2 \right] \nonumber
\\
&\leq  \sum_{t = 1}^T \EXP_{{\set{N}}} \left[ \log (1/\eta)
\mathbbm{1}[i \in {\set{W}}_t] 
\sum_{\ell =1}^L \mathbbm{1}[\ell_t = \ell]
\left(
\mathbbm{1}[X_{i \ell t} = +1] \left|\log \frac{q'_{\ell}}{q_{i\ell}} \right|
+\mathbbm{1}[X_{i \ell t} = -1]  \left| \log \frac{\bar{q}'_{\ell}}{\bar{q}_{i\ell}}\right|
\right) 
\right] \nonumber \\
&= \log (1/\eta) \sum_{\ell = 1}^L \sum_{t =1}^T \mathbbm{1}[i \in \set{W}_t \text{ and } \ell_t =\ell ] \EXP_{\set{N}} \left[
\mathbbm{1}[X_{i \ell t} = +1] \left|\log \frac{q'_{\ell}}{q_{i\ell}} \right|
+\mathbbm{1}[X_{i \ell t} = -1]  \left| \log \frac{\bar{q}'_{\ell}}{\bar{q}_{i\ell}}\right| \right]
\nonumber \\
&= 
\log(1/\eta)  \frac{Tw}{Ln} \sum_{\ell = 1}^L \left( \KL(q_{i'\ell}, q_{i \ell}) + \sqrt{\KL(q_{i'\ell}, q_{i \ell})}\right)\;, 
\label{eq:random-square-term}
\end{align}
where for the last inequality, we used the Pinsker's inequality. Moreover we can compute the expectation of $\sum_{t\neq t'} \set{L}_{t}\set{L}_{t'}$ as follows:
\begin{align*}
&\EXP_{{\set{N}}} \left[ \sum_{t\neq t'} \set{L}_{t}\set{L}_{t'} \right]  = \sum_{t\neq t'}  \EXP_{\set{N}} \left[\set{L}_{t}\set{L}_{t'} \right]
\\
& =  \sum_{t\neq t'}  \EXP_{\set{N}} \left[\mathbbm{1}[i \in \set{W}_t] 
\sum_{\ell =1}^L \mathbbm{1}[\ell_t = \ell]
\left(
\mathbbm{1}[X_{i \ell t} = +1] \log \frac{q'_{\ell}}{q_{i\ell}}
+\mathbbm{1}[X_{i \ell t} = -1]  \log \frac{\bar{q}'_{\ell}}{\bar{q}_{i\ell}}
\right)\right] 
\\ & \cdot \EXP_{\set{N}}\left[\mathbbm{1}[i \in \set{W}_{t'}] 
\sum_{\ell' =1}^L \mathbbm{1}[\ell_{t'} = \ell']
\left(
\mathbbm{1}[X_{i \ell' t'} = +1] \log \frac{q'_{\ell'}}{q_{i\ell'}}
+\mathbbm{1}[X_{i \ell' t'} = -1]  \log \frac{\bar{q}'_{\ell'}}{\bar{q}_{i\ell'}}
\right)\right]
\\
& = \sum_{t \neq t'} \mathbbm{1}[i \in \set{W}_t \text{ and } i \in \set{W}_{t'}] \sum_{\ell =1}^{L} \sum_{\ell' =1}^L 
\mathbbm{1}[\ell_t = \ell \text{ and } \ell_{t'} = \ell'] \KL(q_\ell', q_{i \ell}) \KL(q_{\ell'}', q_{i \ell'})
\\
& = \sum_{\ell =1}^{L} \sum_{\ell' =1}^L T(T-1)\left(\frac{w}{Ln}\right)^2 \KL(q_\ell', q_{i \ell}) \KL(q_{\ell'}', q_{i \ell'})
\\
& = T(T-1)\left(\frac{w}{Ln}\right)^2 \left(  \sum_{\ell =1}^{L} \KL(q_\ell', q_{i \ell})\right) \left(  \sum_{\ell' =1}^L   \KL(q_{\ell'}', q_{i \ell'}) \right)
\\
& =  T(T-1)\left(\frac{w}{Ln}\right)^2 \left(  \sum_{\ell =1}^{L} \KL(q_\ell', q_{i \ell})\right)^2
\\
& = 
\EXP_{\set{N}}[\set{L}]^2 - 
T \left(\frac{w}{Ln}  \right)^2
\left(\sum_{\ell = 1}^L \KL(q'_{\ell}, q_{i \ell})  \right)^2  \;,
\end{align*}
where for the last equality, we use the expression \eqref{eq:random-log-mean} of $\EXP_{\set{N}}[\set{L}]$. Combining \eqref{eq:random-square-term}
 with the above, it follows that:
\begin{align*}
\EXP_{\set{N}} \left[(\set{L} -  \EXP_{\set{N}} [\set{L}]) ^2 \right] = 
\EXP_{\set{N}} [\set{L}^2] - \EXP_{\set{N}} [\set{L}]^2
&\le 
\frac{Tw}{Ln} 
\log(1/\eta) \sum_{\ell = 1}^L \KL(q'_{\ell}, q_{i \ell}) 
+ \sqrt{\KL(q_{i'\ell}, q_{i \ell})}
 \;.
\end{align*} \ep

\section{Proof of Corollary~\ref{cor:lowerbound-uniform}} \label{prf:cor_simple_uniform}

We have:
\begin{align*}
	\set{D}^U_\set{M} (i) & =\min_{h' \in [(h_* + 1)/2, 1]} \KL(h' \updated{p_{21}} + \bar{h}'\updated{\bar{p}_{21}}, q_{i1})
	\\
	& \leq \KL(h_i \updated{p_{21}} + \bar{h}_i \updated{\bar{p}_{21}}, q_{i1})
	\\
	& \stackrel{(a)}{\le} \frac{((h_i \updated{p_{21}} + \bar{h}_i \updated{\bar{p}_{21}})-q_{i1})^2}{q_{i1}(1-q_{i1})}
	\\
	& \stackrel{(b)}{\le} \frac{2}{\eta}  \left( (h_i \updated{p_{11}} + \bar{h}_i \updated{\bar{p}_{11}}) - (h_i \updated{p_{21}} + \bar{h}_i \updated{\bar{p}_{21}}) \right) ^2
	\\
	& = \frac{2}{\eta} \left( \frac{(2h_i - 1) (2 \updated{p_{11}} - 1) +1}{2} - \frac{(2h_i - 1) (2 \updated{p_{21}} - 1)  + 1}{2} \right) ^2
	\\
	& = \frac{2}{\eta}(2h_i - 1)^2 (\updated{p_{11} - p_{21}})^2,
\end{align*}
where $(a)$ stems from the relationship between the KL divergence and the $\chi^2$-divergence and $(b)$ is from (A2). Combining this inequality and Theorem~\ref{thm:lower-random} yield Corollary~\ref{cor:lowerbound-uniform}. \ep

%% file: arxiv.bbl
\begin{thebibliography}{10}

\bibitem{Abbe18}
Emmanuel Abbe.
\newblock Community detection and stochastic block models.
\newblock {\em Foundations and Trends in Communications and Information
  Theory}, 14(1-2):1--162, 2018.

\bibitem{dawid1979maximum}
Alexander~Philip Dawid and Allan~M Skene.
\newblock Maximum likelihood estimation of observer error-rates using the em
  algorithm.
\newblock {\em Journal of the Royal Statistical Society: Series C (Applied
  Statistics)}, 28(1):20--28, 1979.

\bibitem{mallard2024}
Chris F.
\newblock A mallard duck on water.
\newblock \url{https://www.pexels.com/photo/a-mallard-duck-on-water-11798057/},
  2024.
\newblock [Online; accessed 2-February-2024].

\bibitem{gao2016exact}
Chao Gao, Yu~Lu, and Dengyong Zhou.
\newblock Exact exponent in optimal rates for crowdsourcing.
\newblock In {\em Proceedings of the 33nd International Conference on Machine
  Learning}, pages 603--611, 2016.

\bibitem{gao2018community}
Chao Gao, Zongming Ma, Anderson~Y Zhang, Harrison~H Zhou, et~al.
\newblock Community detection in degree-corrected block models.
\newblock {\em The Annals of Statistics}, 46(5):2153--2185, 2018.

\bibitem{gomes2011crowdclustering}
Ryan~G Gomes, Peter Welinder, Andreas Krause, and Pietro Perona.
\newblock Crowdclustering.
\newblock In {\em Advances in neural information processing systems}, pages
  558--566, 2011.

\bibitem{ho2013adaptive}
Chien-Ju Ho, Shahin Jabbari, and Jennifer~Wortman Vaughan.
\newblock Adaptive task assignment for crowdsourced classification.
\newblock In {\em International Conference on Machine Learning}, pages
  534--542, 2013.

\bibitem{hoeffding1963probability}
Wassily Hoeffding.
\newblock Probability inequalities for sums of bounded random variables.
\newblock {\em Journal of the American Statistical Association},
  58(301):13--30, 1963.

\bibitem{geese2024}
Boys in~Bristol~Photography.
\newblock A canada goose grooming while swimming in a lake.
\newblock
  \url{https://www.pexels.com/photo/a-canada-goose-grooming-while-swimming-in-a-lake-7589597/},
  2024.
\newblock [Online; accessed 2-February-2024].

\bibitem{karger2011iterative}
David~R Karger, Sewoong Oh, and Devavrat Shah.
\newblock Iterative learning for reliable crowdsourcing systems.
\newblock In {\em Advances in neural information processing systems}, pages
  1953--1961, 2011.

\bibitem{oh2016}
Ashish Khetan and Sewoong Oh.
\newblock Achieving budget-optimality with adaptive schemes in crowdsourcing.
\newblock In {\em Advances in Neural Information Processing Systems},
  volume~29, 2016.

\bibitem{lai1985asymptotically}
Tze~Leung Lai and Herbert Robbins.
\newblock Asymptotically efficient adaptive allocation rules.
\newblock {\em Advances in applied mathematics}, 6(1):4--22, 1985.

\bibitem{ok2016optimality}
Jungseul Ok, Sewoong Oh, Jinwoo Shin, and Yung Yi.
\newblock Optimality of belief propagation for crowdsourced classification.
\newblock In {\em Proceedings of the 33nd International Conference on Machine
  Learning}, pages 535--544, 2016.

\bibitem{raykar2010learning}
Vikas~C Raykar, Shipeng Yu, Linda~H Zhao, Gerardo~Hermosillo Valadez, Charles
  Florin, Luca Bogoni, and Linda Moy.
\newblock Learning from crowds.
\newblock {\em Journal of Machine Learning Research}, 11(Apr):1297--1322, 2010.

\bibitem{tran2013efficient}
Long Tran-Thanh, Matteo Venanzi, Alex Rogers, and Nicholas~R Jennings.
\newblock Efficient budget allocation with accuracy guarantees for
  crowdsourcing classification tasks.
\newblock In {\em Proceedings of the 2013 international conference on
  Autonomous agents and multi-agent systems}, pages 901--908, 2013.

\bibitem{vinayak2016crowdsourced}
Ramya~Korlakai Vinayak and Babak Hassibi.
\newblock Crowdsourced clustering: Querying edges vs triangles.
\newblock In {\em Advances in Neural Information Processing Systems}, pages
  1316--1324, 2016.

\bibitem{welinder2010multidimensional}
Peter Welinder, Steve Branson, Pietro Perona, and Serge~J Belongie.
\newblock The multidimensional wisdom of crowds.
\newblock In {\em Advances in neural information processing systems}, pages
  2424--2432, 2010.

\bibitem{yun2016optimal}
Se-Young Yun and Alexandre Proutiere.
\newblock Optimal cluster recovery in the labeled stochastic block model.
\newblock In {\em Advances in Neural Information Processing Systems}, pages
  965--973, 2016.

\bibitem{zhang2014spectral}
Yuchen Zhang, Xi~Chen, Dengyong Zhou, and Michael~I Jordan.
\newblock Spectral methods meet em: A provably optimal algorithm for
  crowdsourcing.
\newblock In {\em Advances in neural information processing systems}, pages
  1260--1268, 2014.

\bibitem{zhou2012learning}
Dengyong Zhou, Sumit Basu, Yi~Mao, and John~C Platt.
\newblock Learning from the wisdom of crowds by minimax entropy.
\newblock In {\em Advances in neural information processing systems}, pages
  2195--2203, 2012.

\end{thebibliography}
